\documentclass[10pt,twocolumn,letterpaper]{article}

\usepackage{cvpr}              % To produce the CAMERA-READY version
\usepackage{algorithm}
\usepackage{algpseudocode}
\usepackage{array}
\usepackage{arydshln}
\usepackage{multirow} 
\usepackage{todonotes} 
\usepackage{amsmath}
\DeclareMathOperator*{\argmin}{arg\,min}
\newcommand{\ours}[0]{\texttt{MirrorCheck}}
\definecolor{cvprblue}{rgb}{0.21,0.49,0.74}
\usepackage[pagebackref,breaklinks,colorlinks,allcolors=cvprblue]{hyperref}

\def\paperID{30516} % *** Enter the Paper ID here
\def\confName{CVPR}
\def\confYear{2026}

\title{\ours{}: Efficient Adversarial Defense for Vision-Language Models}

\author{
Samar Fares\textsuperscript{1}\thanks{Equal contribution} \quad
Klea Ziu\textsuperscript{1}\footnotemark[1] \quad
Toluwani Aremu\textsuperscript{1}\footnotemark[1] \quad
Nikita Durasov\textsuperscript{2} \quad
Martin Takáč\textsuperscript{1} \\
Pascal Fua\textsuperscript{3} \quad
Ivan Laptev\textsuperscript{1} \quad
Karthik Nandakumar\textsuperscript{1,4}\\
\textsuperscript{\tiny{1}}\small{Mohamed Bin Zayed University of Artificial Intelligence (MBZUAI)} \\
\textsuperscript{\tiny{2}}\small{NVIDIA} \quad
\textsuperscript{\tiny{3}}\small{École Polytechnique Fédérale de Lausanne (EPFL)} \\
\textsuperscript{\tiny{4}}\small{Michigan State University}
}

\begin{document}
\maketitle
\begin{abstract}
Vision-Language Models (VLMs) are increasingly susceptible to sophisticated adversarial attacks, including adaptive strategies specifically designed to bypass existing defenses. To address this vulnerability, we propose \ours{}, a robust and model-agnostic detection framework that operates effectively in both unimodal and multimodal settings. \ours{} leverages Text-to-Image (T2I) models to regenerate visual content from captions produced by the target model and assesses semantic consistency by comparing feature-space embeddings between the original and synthesized images. To enhance robustness against adaptive attacks, \ours{} introduces a stochastic defense strategy that randomly selects T2I generators and image encoders from a diverse model zoo. Additionally, we incorporate a novel One-Time-Use (OTU) perturbation applied to the selected encoder embeddings, regulated by a scaling factor, which decreases the effectiveness of adaptive attacks. Extensive experiments across multiple threat scenarios demonstrate that \ours{} consistently outperforms baseline methods, and maintains its utility even under strong adaptive adversarial conditions. 
\end{abstract}

\vspace{-1em}    
\section{Introduction}
\label{sec:intro}

Vision-Language Models (VLMs) have emerged as powerful tools at the intersection of computer vision (CV) and natural language processing (NLP), enabling machines to reason jointly across modalities and deliver state-of-the-art performance in tasks such as image captioning (IC), visual question answering (VQA), and image text retrieval \citep{bao2022all, li2022blip, li2023blip, zhu2023minigpt, li-etal-2023-lavis}. However, alongside their impressive capabilities comes an increased susceptibility to adversarial attacks, maliciously crafted inputs that cause models to produce incorrect or misleading outputs with imperceptible perturbations \citep{yin2023vlattack, zhao2023evaluate, vemprala2023chatgpt}. \vspace{0.5em}

\begin{figure*}[t]
    \centering
    \includegraphics[width=1\linewidth]{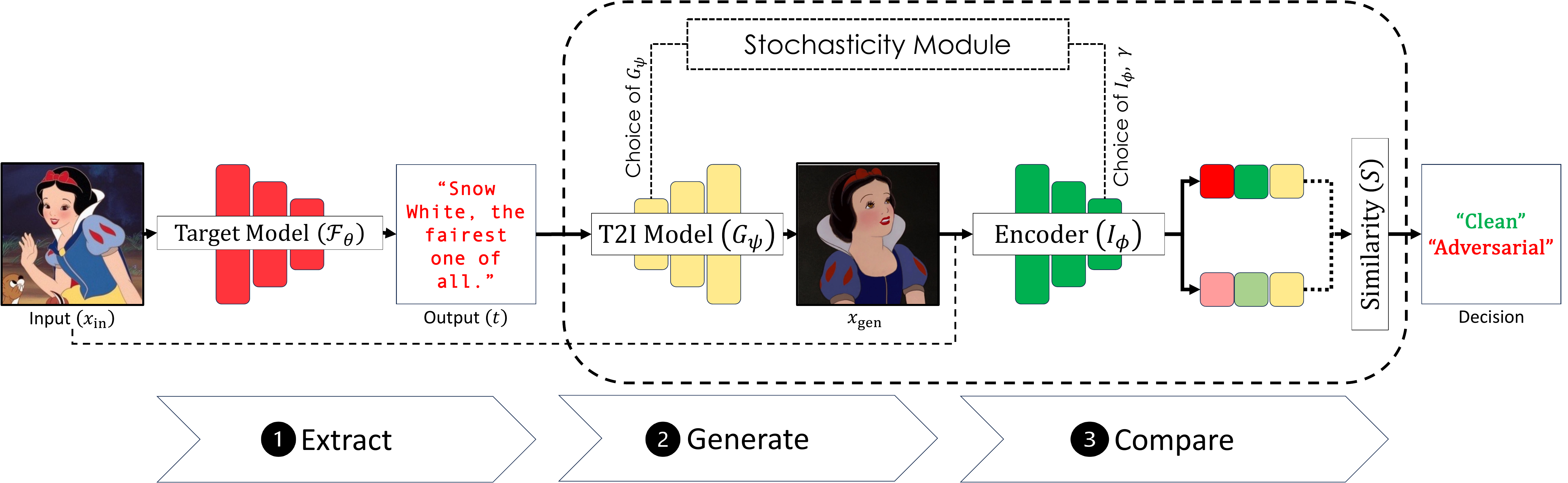}
    \caption{\textbf{\ours{} framework:} (1) \textbf{Extract}: An input image is passed through the victim model to generate an output (caption/description/answer/classification). (2) \textbf{Generate}: The output is fed to a randomly selected T2I diffusion model, returning a generated image $x_{gen}$. (3) \textbf{Compare}: We extract and compare feature embeddings from both original $x_{in}$ and generated $x_{gen}$ images using randomly selected and uniquely perturbed image encoders. Significant embedding discrepancies indicate potential adversarial attacks.}
    \label{fig:otupipeline}
\end{figure*}

\noindent Various strategies, such as detectors \citep{metzen2017detecting, odds}, purifiers \citep{DiffPure, defensegan}, adversarial training \citep{madry2018towards}, and certified defenses \citep{certrandsmooth}, have been proposed to defend against adversarial threats. However, these methods are specifically tailored for image classification tasks, sometimes requiring expensive retraining and task-specific tuning, while remaining vulnerable to adaptive attacks \citep{ObfuscatedGG}. While recent efforts \citep{wang2024pretrainedmodelguidedfinetuning, mao2023understandingzeroshotadversarialrobustness, xie2024gradsafe, zhang2024pip} have explored improving the adversarial robustness of VLMs, these approaches do not provide empirical guarantees against white-box adaptive attacks. \vspace{0.5em}

\noindent To address this gap, we introduce \ours{}, a novel, model-agnostic adversarial detection framework for VLMs. Specifically, we leverage Text-to-Image (T2I) models to regenerate images from captions produced by the potentially compromised model, and compare the original and generated images in the embedding space using randomly chosen image encoders. A lower similarity score indicates a likely adversarial sample. We propose two variants of our method: \textbf{Vanilla}, which establishes the core detection pipeline using T2I generation and embedding comparison; and \textbf{Stochastic}, which leverages randomized model choice and weight transformations for robustness against adaptive attacks. This layered stochasticity increases the search space for attackers, rendering white-box adaptive attacks computationally intractable. To summarize: \textbf{(i)} We present \ours{}, a framework for detecting adversarial samples in VLMs. \ours{} is a plug-and-play approach which doesn't require training and is model-agnostic. \textbf{(ii)} We further propose a stochastic extension of \ours{} that introduces randomness and controlled perturbations to thwart adaptive attacks. \textbf{(iii)} Extensive empirical evaluations across various attack settings demonstrates that \ours{} outperforms baselines and maintains strong performance under adaptive threat models. Results also reveals that \ours{} can also generalize to unimodal tasks.
\section{Related Work}
\label{sec:background}

In this section, we briefly review a few attacks and defenses relevant to our study.

\subsection{Adversarial Attacks}
Adversarial attacks exploit model vulnerabilities through perturbations that cause misclassification or targeted misbehavior. Early research focused on unimodal architectures, particularly CNNs for image classification \cite{goodfellow2015explaining, kurakin2016adversarial, madry2018towards, carlini2017towards}. These attacks are categorized into white-box settings (providing full model access and enabling gradient-based methods like FGSM \cite{goodfellow2015explaining} and PGD \cite{madry2018towards}) and black-box settings (relying on transferability or query-based methods \cite{EOT}). Recent advances have extended attacks to multimodal systems, particularly vision-language models (VLMs). AttackVLM \cite{zhao2023evaluate} introduces transfer-based and query-based strategies targeting VLMs in black-box scenarios, while VLATTACK \cite{yin2023vlattack} and Attack-Bard \cite{bard} combine image and text perturbations. These attacks exploit the architectural complexity of VLMs, where perturbations can impact both visual and textual modalities, potentially making them more vulnerable than their unimodal counterparts \cite{cao2022tasa, kovatchev2022longhorns, zhang2022towards, aafaq2021controlled}.

\subsection{Adversarial Defenses}
Traditional adversarial defenses for unimodal tasks include detection methods \cite{odds, FeatureSqueeze, magnet}, purification techniques \cite{DiffPure, defensegan, puvae}, adversarial training \cite{adversarialtraining, madry2018towards}, and certified defenses \cite{certrandsmooth, denoisesmooth}. However, these approaches face critical limitations when applied to VLMs, as they operate on single modalities and cannot account for complex visual-linguistic interactions that adversaries exploit. Furthermore, conventional defenses are vulnerable to adaptive attacks where adversaries with white-box access systematically bypass protection mechanisms \cite{ObfuscatedGG, tramer2020adaptive, carlini2016defensive}.\vspace{0.5em}

\noindent More recently, a new paradigm of training-free defense strategies \cite{zhou2024defending, xie2024gradsafe, zhang2024pip, sun2024safeguarding} have been proposed to safeguard VLMs, validating the paradigm's effectiveness. Our work follows this paradigm and introduces a framework that leverages T2I models and employs stochasticity to resist adaptive attacks. We provide detailed descriptions of specific attack methods and baseline defenses in \cref{lrbck}.
\section{Method}

\noindent Let  $\mathcal{F}_\theta(x_{\text{in}}; p) \rightarrow t$ be the victim model which can be a VLM or a Classification model (any models that generates a description text \( t \)  in response to an input image), where $x_{\text{in}}$ is the input image which may be clean ($x_{\text{clean}})$ or adversarial ($x_{\text{adv}}$), $p$ is the input prompt, and $t$ is the resulting output text. In certain tasks, such as image captioning or text retrieval, the input prompt $p$ may remain empty. Let
$\mathcal{I}_{\phi}(x) \rightarrow z$ be a pretrained image encoder and let $G_{\psi}(t) \rightarrow x_{\text{gen}}$ denote a pretrained text-conditioned image generation model producing image $x_{\text{gen}}$.
%is the Text2Img model used by the defender to generate the image $x_{\text{gen}}$ to be used for comparison.

% \textcolor{red}{We need to clearly define the type of task the VLMs we are targeting perform — whether it is limited to captioning or extends to other tasks. We should also check whether attacks exist for tasks beyond captioning. Additionally, we must explicitly state the types of attacks considered in our threat model (white-box vs. black-box, targeted vs. untargeted).}

\subsection{Threat Model}

\noindent Our framework is designed to detect adversarial attacks, irrespective of the attacker's level of knowledge. In this scenario, there are two parties: 

\vspace{-1em}
\paragraph{Attacker.}  
The attacker's goal is to generate an adversarial image \( x_{\text{adv}} = x_{\text{clean}} + \delta \) that causes the victim model to produce an incorrect caption or classification. The attack can be \textbf{targeted}, where the generated text \( t \) matches a predefined adversarial target, or \textbf{untargeted}, where the model is simply forced to misinterpret or misdescribe the input image. In both cases, the perturbation \( \delta \) is constrained within an \( \ell \)-norm bounded adversarial budget to ensure minimal perceptibility while maximizing the likelihood of deception. We make no assumptions about the adversary’s level of access to the victim model, they may have full knowledge of its architecture, parameters, and training data, or they may operate in a black-box setting with no such information.

\paragraph{Defender.}  
The defender aims to correctly classify input images as either \textit{clean} or \textit{adversarial} by assessing the consistency between the model’s interpretation of the input and a reference image generated from the model’s textual output. The detection mechanism does not rely on knowledge of the specific adversarial attack strategy and assumes only black-box access to the victim model. Furthermore, the defender does not have access to any ground-truth clean reference image, only the input image \( x_{\text{in}} \), which may be either clean or adversarial.

% #####################################################################################################################################
\begin{table*}[t]
\caption{\small{\textbf{Similarity scores between original and regenerated images using \texttt{Stochastic MirrorCheck}}. The tasks used are image captioning (IC), image description (ID), visual question answering (VQA), and image classification (CL).
Clean images consistently achieve high similarity scores, while adversarial examples show degraded similarity, enabling effective detection across models and attack types. Results shown for random CLIP encoder selection with One-Time-Use perturbations across different ensemble sizes and noise scales.}}
% \vspace{-1.5em}
\label{SD-similarity}
\begin{center}
     \setlength\tabcolsep{8pt}
\resizebox{\linewidth}{!}{
\begin{tabular}{lllccccccccccccccc} 
\toprule
\multirow{3}{*}{Victim Model} &  \multirow{3}{*}{Task} & \multirow{3}{*}{Attack Setting} & \multicolumn{15}{c}{CLIP Image Encoders (Random Selection + Noise)}  \\ 
\cmidrule(lr){4-18}
& & & \multicolumn{3}{c}{1 Encoder} & \multicolumn{3}{c}{3 Encoders} & \multicolumn{3}{c}{5 Encoders} & \multicolumn{3}{c}{7 Encoders} & \multicolumn{3}{c}{10 Encoders}  \\ 
\cmidrule(lr){4-6} \cmidrule(lr){7-9} \cmidrule(lr){10-12} \cmidrule(lr){13-15} \cmidrule(lr){16-18} &
& & 5e-6 & 5e-4 & 1e-3 & 5e-6 & 5e-4 & 1e-3 & 5e-6 & 5e-4 & 1e-3 & 5e-6 & 5e-4 & 1e-3 & 5e-6 & 5e-4 & 1e-3    \\ 
\midrule
\multirow{3}{*}{UniDiffuser} &\multirow{3}{*}{IC} &Clean&
        0.721 & 0.624 & 0.740 & 0.651 & 0.701 & 0.685 & 0.694 & 0.647 & 0.715 & 0.648 & 0.693 & 0.670 & 0.665 & 0.670 & 0.692
\\ 
\cdashline{3-18}
& &AttackVLM-T &
        0.502 & 0.341 & 0.568 & 0.370 & 0.501 & 0.477 & 0.494 & 0.399 & 0.549 & 0.402 & 0.472 & 0.458 & 0.424 & 0.441 & 0.494 \\ 
\cdashline{3-18}
 & & AttackVLM-Q &
        0.498 & 0.294 & 0.542 & 0.336 & 0.448 & 0.395 & 0.424 & 0.332 & 0.446 & 0.340 & 0.411 & 0.375 & 0.366 & 0.372 & 0.408 \\ 
\midrule

\multirow{3}{*}{BLIP}  & \multirow{3}{*}{IC} & Clean&
        0.707 & 0.610 & 0.730 & 0.633 & 0.686 & 0.672 & 0.676 & 0.628 & 0.700 & 0.629 & 0.675 & 0.652 & 0.647 & 0.653 & 0.676
\\ 
\cdashline{3-18}
&& AttackVLM-T &
        0.481 & 0.323 & 0.547 & 0.349 & 0.450 & 0.454 & 0.419& 0.362 & 0.480 & 0.353 & 0.423 & 0.412 & 0.375 & 0.391 & 0.448\\ \cdashline{3-18}  
&& AttackVLM-Q &
        0.508 & 0.299 & 0.555 & 0.350 & 0.460 & 0.460 & 0.436 & 0.346 & 0.407 & 0.354 & 0.424 & 0.389 & 0.379 & 0.384 & 0.421
 \\ 
\midrule

\multirow{4}{*}{BLIP-2} &\multirow{3}{*}{IC}&Clean&
        0.729 & 0.636 & 0.744 & 0.655 & 0.705 & 0.687 & 0.697 & 0.664 & 0.718 & 0.651 & 0.695 & 0.684 & 0.668 & 0.675 & 0.695
\\ 
\cdashline{3-18}
&& AttackVLM-T &
        0.504 & 0.345 & 0.563 & 0.376 & 0.473 & 0.475 & 0.443 & 0.381 & 0.503 & 0.377 & 0.447 & 0.434 & 0.399 & 0.413 & 0.467\\ 
        \cdashline{3-18}  
&& AttackVLM-Q &
        0.380 & 0.323 & 0.484 & 0.343 & 0.352 & 0.408 & 0.382 & 0.345 & 0.401 & 0.340 & 0.388 & 0.395 & 0.372 & 0.388 & 0.421 \\ 
        \cdashline{3-18}  
&ID& Attack-Bard &
        0.484 & 0.422 & 0.536 & 0.379 & 0.444 & 0.498 & 0.399 & 0.416 & 0.468 & 0.377 & 0.451 & 0.427 & 0.399 & 0.420 & 0.461 \\ 
        \midrule

\multirow{3}{*}{Img2Prompt} & \multirow{3}{*}{VQA}
&Clean&
        0.675 & 0.563 & 0.705 & 0.589 & 0.652 & 0.637 & 0.637& 0.585 & 0.677 & 0.586& 0.638 & 0.616& 0.605 & 0.613 & 0.642
\\ 
\cdashline{3-18}
&& AttackVLM-T &
        0.482 & 0.317 & 0.547 & 0.345 & 0.449 & 0.455 & 0.416 & 0.359 & 0.479 & 0.349 & 0.422 & 0.412 & 0.372 & 0.388 & 0.477\\ 
    \cdashline{3-18}
 && AttackVLM-Q &
        0.517 & 0.309 & 0.561 & 0.361 & 0.470 & 0.467 & 0.447 & 0.356 & 0.414 & 0.365 & 0.431& 0.396 & 0.390 & 0.392 & 0.427 \\ 
        \midrule

\multirow{2}{*}{LLaVA }& \multirow{2}{*}{VQA} &Clean&
        0.680 & 0.823 & 0.755 & 0.733 & 0.714 & 0.741 & 0.728 & 0.810 & 0.748 & 0.725 & 0.706 & 0.733 & 0.712 & 0.798 & 0.742
\\ 
\cdashline{3-18}
&& {Attack-MMFM} &
        0.539 & 0.724 & 0.626 & 0.599 & 0.596 & 0.617 & 0.618 & 0.710 & 0.641 & 0.608 & 0.602 & 0.625 & 0.595 & 0.695 & 0.632 \\ 
        \midrule

\multirow{2}{*}{OpenFlamingo }  & \multirow{2}{*}{VQA} &Clean&
        0.690 & 0.817 & 0.756 & 0.728 & 0.723 & 0.743 & 0.734 & 0.804 & 0.749 & 0.720 & 0.715 & 0.735 & 0.708 & 0.791 & 0.742
\\ 
\cdashline{3-18}
&& {Attack-MMFM} &
        0.535 & 0.714 & 0.618 & 0.584 & 0.609 & 0.612 & 0.609 & 0.701 & 0.635 & 0.596 & 0.614 & 0.620 & 0.582 & 0.688 & 0.625 \\ 
        \midrule

\multirow{2}{*}{MiniGPT-4 }  &\multirow{2}{*}{VQA} &Clean&
        0.651 & 0.536 & 0.684 & 0.561 & 0.628 & 0.618 & 0.612 & 0.560 & 0.646 & 0.559 & 0.613 & 0.593 & 0.578 & 0.587 & 0.620
\\ 
\cdashline{3-18}
&& AttackVLM-T &
        0.568 & 0.457 & 0.620 & 0.472 & 0.548 & 0.551 & 0.523 & 0.481 & 0.576 & 0.469 & 0.532 & 0.519 & 0.489 & 0.504 & 0.549 \\ 
         \midrule

\multirow{6}{*}{DenseNet} &\multirow{6}{*}{CL} &Clean&
        0.543 & 0.740 & 0.705 & 0.671 & 0.674 & 0.667 & 0.692 & 0.658 & 0.695 & 0.665 & 0.688 & 0.671 & 0.652 & 0.660 & 0.679
\\ 
\cdashline{3-18}
&& FGSM &
        0.444 & 0.666 & 0.572 & 0.537 & 0.548 & 0.553 & 0.579 & 0.521 & 0.584 & 0.535 & 0.572 & 0.558 & 0.518 & 0.541 & 0.567
\\ 
\cdashline{3-18}
&& BIM &
        0.507 & 0.713 & 0.593 & 0.554 & 0.532 & 0.579 & 0.601 & 0.542 & 0.598 & 0.548 & 0.586 & 0.571 & 0.531 & 0.553 & 0.581
        \\       
        \cdashline{3-18}
&& PGD &
        0.495 & 0.705 & 0.585 & 0.546 & 0.524 & 0.571 & 0.593 & 0.534 & 0.590 & 0.540 & 0.578 & 0.563 & 0.523 & 0.545 & 0.573
        \\         
        \cdashline{3-18}
&& DeepFool &
        0.475 & 0.690 & 0.565 & 0.525 & 0.510 & 0.555 & 0.575 & 0.515 & 0.570 & 0.520 & 0.560 & 0.545 & 0.505 & 0.525 & 0.555
        \\    
                \cdashline{3-18}
&& C\&W  &
        0.460 & 0.680 & 0.555 & 0.515 & 0.500 & 0.545 & 0.565 & 0.505 & 0.560 & 0.510 & 0.550 & 0.535 & 0.495 & 0.515 & 0.545
        \\ 
         \midrule

\multirow{6}{*}{MobileNet} &\multirow{6}{*}{CL} &Clean&
        0.668 & 0.790 & 0.729 & 0.704 & 0.705 & 0.719 & 0.745 & 0.698 & 0.738 & 0.702 & 0.726 & 0.712 & 0.688 & 0.695 & 0.721
\\ 
\cdashline{3-18}
&& FGSM &
        0.520 & 0.712 & 0.606 & 0.612 & 0.585 & 0.607 & 0.635 & 0.598 & 0.629 & 0.605 & 0.618 & 0.610 & 0.585 & 0.592 & 0.615
\\ 
\cdashline{3-18}
&& BIM &
        0.503 & 0.693 & 0.581 & 0.565 & 0.538 & 0.576 & 0.605 & 0.572 & 0.598 & 0.575 & 0.590 & 0.582 & 0.558 & 0.565 & 0.585
        \\       
        \cdashline{3-18}
&& PGD &
        0.495 & 0.685 & 0.573 & 0.557 & 0.530 & 0.568 & 0.597 & 0.564 & 0.590 & 0.567 & 0.582 & 0.574 & 0.550 & 0.557 & 0.577
        \\         
        \cdashline{3-18}
&& DeepFool &
        0.475 & 0.670 & 0.555 & 0.540 & 0.515 & 0.552 & 0.580 & 0.548 & 0.573 & 0.550 & 0.565 & 0.557 & 0.535 & 0.542 & 0.562
        \\    
                \cdashline{3-18}
&& C\&W &
        0.465 & 0.660 & 0.545 & 0.530 & 0.505 & 0.542 & 0.570 & 0.538 & 0.563 & 0.540 & 0.555 & 0.547 & 0.525 & 0.532 & 0.552
        \\ 
        
        \bottomrule
\end{tabular}
}
\end{center}
\end{table*}

% #####################################################################################################################################

% #####################################################################################################################################

\begin{table*}[t]
\caption{\small{\textbf{Detection accuracy of \texttt{Stochastic MirrorCheck} across diverse victim models and attack types.} The method achieves consistently high detection rates (65-99\%) across VLM attacks (AttackVLM, Attack-Bard, Attack-MMFM) and classification attacks (FGSM, BIM, PGD, DeepFool, C\&W), demonstrating robust performance with randomized encoder selection and One-Time-Use perturbations.}}
% \vspace{-1.5em}
\label{SD-detection}
\begin{center}
     \setlength\tabcolsep{8pt}
\resizebox{\linewidth}{!}{
\begin{tabular}{llccccccccccccccc} 
\toprule
\multirow{3}{*}{Victim Model} & \multirow{3}{*}{Setting} & \multicolumn{15}{c}{CLIP Image Encoders (Random Selection + Noise)}  \\ 
\cmidrule(lr){3-17}
& & \multicolumn{3}{c}{1 Encoder} & \multicolumn{3}{c}{3 Encoders} & \multicolumn{3}{c}{5 Encoders} & \multicolumn{3}{c}{7 Encoders} & \multicolumn{3}{c}{10 Encoders}  \\ 
\cmidrule(lr){3-5} \cmidrule(lr){6-8} \cmidrule(lr){9-11} \cmidrule(lr){12-14} \cmidrule(lr){15-17}
& & 5e-6 & 5e-4 & 1e-3 & 5e-6 & 5e-4 & 1e-3 & 5e-6 & 5e-4 & 1e-3 & 5e-6 & 5e-4 & 1e-3 & 5e-6 & 5e-4 & 1e-3    \\ 
\midrule
\multirow{2}{*}{UniDiffuser}& AttackVLM-T &
        0.913 & 0.895 & 0.903 & 0.925 & 0.943 & 0.858 & 0.933 & 0.890 & 0.910 & 0.908 & 0.912 & 0.915 & 0.920 & 0.918 & 0.917 \\ 
\cdashline{2-17}
 & AttackVLM-Q &
        0.955 & 0.902 & 0.968 & 0.952 & 0.968 & 0.937 & 0.973 & 0.975 & 0.980 & 0.977 & 0.978 & 0.980 & 0.975 & 0.973 & 0.992 \\ 
\midrule
\multirow{3}{*}{BLIP} & AttackVLM-T & 
        0.905 & 0.908 & 0.908 & 0.918 & 0.932 & 0.903 & 0.922 & 0.915 & 0.925 & 0.930 & 0.930 & 0.918 & 0.930 & 0.923 & 0.925 \\ \cdashline{2-17}  
& AttackVLM-Q &
        0.928 & 0.887 & 0.943 & 0.913 & 0.937 & 0.915 & 0.958 & 0.943 & 0.963 & 0.962 & 0.947 & 0.955 & 0.965 & 0.945 & 0.955 
 \\ 
\midrule
\multirow{3}{*}{BLIP-2} & AttackVLM-T &
        0.912 & 0.892 & 0.912 & 0.920 & 0.932 & 0.907 & 0.923& 0.923 & 0.928 & 0.927 & 0.935 & 0.932 & 0.930 & 0.927 & 0.938 \\ 
        \cdashline{2-17}  
& AttackVLM-Q &
        0.945 & 0.903 & 0.967 & 0.948 & 0.957 & 0.932 & 0.975 & 0.97 & 0.985 & 0.982 & 0.978 & 0.977 & 0.978 & 0.970 & 0.990 \\ 
        \cdashline{2-17}  
& Attack-Bard &
        0.883 & 0.790 & 0.827 & 0.890 & 0.873 & 0.900 & 0.903 & 0.952 & 0.903& 0.942 & 0.913 & 0.902 & 0.927 & 0.920& 0.938 \\ 
        \midrule
\multirow{2}{*}{Img2Prompt} & AttackVLM-T & 
        0.848 & 0.840 & 0.843 & 0.878 & 0.873 & 0.853 & 0.882 & 0.860& 0.867 & 0.878 & 0.875 & 0.883 & 0.895 & 0.878 & 0.882 \\ 
    \cdashline{2-17}
 & AttackVLM-Q &
        0.880 & 0.806 & 0.907 & 0.861 & 0.863 & 0.855 & 0.886 & 0.857 & 0.905 & 0.870 & 0.877 & 0.905 & 0.887 & 0.882 & 0.920 \\ 
        \midrule
{LLaVA}  & {Attack-MMFM} &
        0.788 & 0.728 & 0.733 & 0.812 & 0.762 & 0.738 & 0.818 & 0.812 & 0.783 & 0.832 & 0.815 & 0.812 & 0.845 & 0.833 & 0.820 \\ 
        \midrule
{OpenFlamingo}  & {Attack-MMFM} &
        0.800 & 0.740 & 0.765 & 0.777 & 0.750 & 0.750 & 0.807 & 0.785 & 0.767 & 0.800 & 0.780 & 0.797 & 0.797 & 0.793 & 0.785 \\ 
        \midrule
MiniGPT-4 & AttackVLM-T &
        0.642 & 0.623 & 0.632 & 0.660 & 0.660 & 0.642 & 0.655 & 0.665 & 0.667 & 0.655 & 0.665 & 0.657 & 0.655 & 0.665 & 0.655 \\ 
        \midrule
\multirow{6}{*}{DenseNet} & FGSM &
        0.850 & 0.840 & 0.800 & 0.845 & 0.835 & 0.795 & 0.852 & 0.842 & 0.802 & 0.847 & 0.837 & 0.798 & 0.849 & 0.839 & 0.801 \\
\cdashline{2-17}
& BIM &
        0.860 & 0.830 & 0.830 & 0.855 & 0.825 & 0.825 & 0.862 & 0.832 & 0.832 & 0.857 & 0.827 & 0.828 & 0.859 & 0.829 & 0.831 \\
\cdashline{2-17}
& PGD &
        0.845 & 0.825 & 0.815 & 0.840 & 0.820 & 0.810 & 0.847 & 0.827 & 0.817 & 0.842 & 0.822 & 0.812 & 0.844 & 0.824 & 0.814 \\
\cdashline{2-17}
& DeepFool &
        0.835 & 0.815 & 0.795 & 0.830 & 0.810 & 0.790 & 0.837 & 0.817 & 0.797 & 0.832 & 0.812 & 0.792 & 0.834 & 0.814 & 0.794 \\
\cdashline{2-17}
& C\&W &
        0.825 & 0.805 & 0.785 & 0.820 & 0.800 & 0.780 & 0.827 & 0.807 & 0.787 & 0.822 & 0.802 & 0.782 & 0.824 & 0.804 & 0.784 \\
\midrule
\multirow{6}{*}{MobileNet} & FGSM (0.3) &
        0.780 & 0.770 & 0.750 & 0.775 & 0.765 & 0.745 & 0.782 & 0.772 & 0.752 & 0.777 & 0.767 & 0.747 & 0.779 & 0.769 & 0.749 \\
\cdashline{2-17}
& FGSM (0.1) &
        0.850 & 0.850 & 0.790 & 0.845 & 0.845 & 0.785 & 0.852 & 0.852 & 0.792 & 0.847 & 0.847 & 0.787 & 0.849 & 0.849 & 0.789 \\
\cdashline{2-17}
& BIM &
        0.790 & 0.790 & 0.780 & 0.785 & 0.785 & 0.775 & 0.792 & 0.792 & 0.782 & 0.787 & 0.787 & 0.777 & 0.789 & 0.789 & 0.779 \\
\cdashline{2-17}
& PGD &
        0.775 & 0.775 & 0.765 & 0.770 & 0.770 & 0.760 & 0.777 & 0.777 & 0.767 & 0.772 & 0.772 & 0.762 & 0.774 & 0.774 & 0.764 \\
\cdashline{2-17}
& DeepFool &
        0.760 & 0.750 & 0.740 & 0.755 & 0.745 & 0.735 & 0.762 & 0.752 & 0.742 & 0.757 & 0.747 & 0.737 & 0.759 & 0.749 & 0.739 \\
\cdashline{2-17}
& C\&W &
        0.745 & 0.735 & 0.725 & 0.740 & 0.730 & 0.720 & 0.747 & 0.737 & 0.727 & 0.742 & 0.732 & 0.722 & 0.744 & 0.734 & 0.724 \\
\bottomrule
\end{tabular}
}
\end{center}
\end{table*}

% #####################################################################################################################################

\subsection{\texttt{MirrorCheck} Pipeline}

\subsubsection{\texttt{Vanilla MirrorCheck.}}

\noindent The framework is illustrated in \cref{fig:otupipeline}. The key observation lies in the deviation of text generated by adversarial images from the content of the input image, which is the primary objective of the attack. \vspace{0.5em}

\noindent Given an input image \( x_{\text{in}} \), we first obtain a textual or a class description using the victim model:
\begin{equation}
    t = \mathcal{F}_\theta(x_{\text{in}}, p).
\end{equation}
This text is then used as input to a pretrained text-to-image model \( G_\psi \), which generates a reconstructed image:
\begin{equation}
    x_{\text{gen}} = G_\psi(t).
\end{equation}
If the input image is clean, the generated image \( x_{\text{gen}} \) should preserve semantic consistency with \( x_{\text{in}} \). However, if \( x_{\text{in}} \) has been adversarially altered, the perturbation may distort the semantic information, leading to a discrepancy between \( x_{\text{in}} \) and \( x_{\text{gen}} \). To quantify this discrepancy, we compare their feature embeddings obtained as follows: 
\begin{equation}
    z_{in} = \mathcal{I}_{\phi}(x_{\text{in}}), \quad z_{\text{gen}} = \mathcal{I}_{\phi}(x_{\text{gen}}).
\end{equation}
Subsequently, we employ an adversarial detector $\mathcal{D}(x) \rightarrow [0,1]$, which categorizes the image into either the "adversarial" class ($1$) or the "clean" class ($0$) based on the similarity between the embeddings, with $\tau$ serving as the decision threshold parameter, i.e.
\[
\mathcal{D}(x)=  
\begin{cases}
    1, \quad & \text{if } S(  z_{in} , z_{\text{gen}}) < \tau,\\
    0,              & \text{otherwise}
\end{cases}.
\]
Where $S$ is the similarity metric between these embeddings, we employ the cosine similarity:
\begin{equation}
    S(z_{in}, z_{\text{gen}} ) = \frac{z_{in} \cdot z_{\text{gen}} }{\|z_{in}\| \|z_{\text{gen}} \|}.
\end{equation}
The optimal value of \(\tau\) is determined using the Receiver Operating Characteristic (ROC) curve analysis. Specifically, we identify the point on the ROC curve where the difference between the true positive rate TPR  (the proportion of actual adversarial images correctly
identified) and the false positive rate FPR  (the proportion of clean images
incorrectly flagged as adversarial) is maximized. This approach ensures a balanced trade-off between detection sensitivity and robustness, making \(\tau\) an effective decision threshold for identifying adversarial samples. However, the choice of \(\tau\) may vary based on the characteristics of the specific text-to-image models or pretrained image encoders used, and we recommend calibrating \(\tau\) accordingly to account for variations in model behavior.
\vspace{-1em}
\paragraph{Intuition behind image-image similarity.} Instead of directly comparing $x_{\text{in}}$ (the input image) with the generated caption $t$, we opted to calculate the similarity between $x_{\text{in}}$ and $x_{\text{gen}}$ (the newly generated image). This decision is based on evidence in the literature indicating that these models struggle with positional relationships and variations in verb usage within sentences. This suggests that VLMs may function more like bags-of-words and, consequently, which could limit their reliability for optimizing cross-modality similarity \cite{yuksekgonul2022and}. Furthermore, we selected this embedding-based similarity metric over conventional metrics like SSIM or FID because those methods may fail to capture semantic equivalence in cases where the T2I model generates a visually different image that is still semantically similar. By utilizing vector embeddings, we aim to maintain high similarity scores in such scenarios, ensuring robustness and reliability even when T2I outputs exhibit variability in their visual representation. Recognizing the potential issue introduced by a single image encoder used for similarity assessment (i.e., if it was used to generate the adversarial samples), the defender employs an ensemble of pretrained image encoders. The final similarity score is obtained by averaging across $n$ predetermined encoders:
\begin{equation}
    S_{\text{ensemble}} = \frac{1}{n} \sum_{k=1}^{n} S(z_{{in}_k}, z_{\text{gen}_k}).
\end{equation}

%  The similarity between these embeddings is measured using cosine similarity:
% \begin{equation}
%     S(z_{i_k}, z_{\text{gen}, i_k}) = \frac{z_{i_k} \cdot z_{\text{gen}, i_k}}{\|z_{i_k}\| \|z_{\text{gen}, i_k}\|}.
% \end{equation}
% The final similarity score is obtained by averaging across all selected encoders:
% \begin{equation}
%     S_{\text{ensemble}} = \frac{1}{n} \sum_{k=1}^{n} S(z_{i_k}, z_{\text{gen}, i_k}).
% \end{equation}

\subsubsection{\texttt{Stochastic MirrorCheck.}}
We extend \texttt{Vanilla MirrorCheck} through a comprehensive stochastic defense paradigm. While \texttt{Vanilla MirrorCheck} employs fixed T2I models and predetermined image encoders (single or ensemble), \texttt{Stochastic MirrorCheck} introduces three key innovations that enhance robustness against adaptive attacks: Randomized T2I model selection, Stochastic encoder deployment, and  One-Time-Use (OTU) perturbations. This randomization exponentially increases the computational complexity of mounting successful adaptive attacks. It requires adversaries to simultaneously predict \textbf{which specific models will be deployed} and \textbf{precisely how their parameters will be perturbed}. Formally, the defender maintains a set of $M$ pretrained text-conditioned-image generation models: 
\begin{equation}
    G = \{ G_{\psi_1}, G_{\psi_2}, ..., G_{\psi_M} \}.
\end{equation}
where $x_{\text{gen}}$ is generated by a one randomly selected generation model. Additionally, the defender maintains a model zoo of \( N \) pretrained image encoders. At detection time, a subset of \( n \) encoders is randomly selected:
\begin{equation}
    \mathcal{I}_{\text{subset}} = \{  \mathcal{I}_{\phi_{1}},  \mathcal{I}_{\phi_{2}}, ...,  \mathcal{I}_{\phi_{n}} \}, \quad n \ll N.
\end{equation}
To further enhance robustness, each selected encoder undergoes \texttt{OTU} perturbation, where its parameters are modified by injecting small random noise \( \gamma \):
\begin{equation}
    \hat{\phi}_{k} = \phi_{k} + \gamma, \quad \forall k \in \mathcal{I}_{\text{subset}}.
\end{equation}
This ensures that the attacker cannot reliably optimize against a fixed set of encoders.
The perturbed encoders extract feature embeddings from both the original and regenerated images:
\begin{equation}
    z_{i_k} = \mathcal{I}_{\hat{\phi}_{i_k}}(x_{\text{in}}), \quad z_{\text{gen}, i_k} = \mathcal{I}_{\hat{\phi}_{i_k}}(x_{\text{gen}}).
\end{equation}
Finally, \( S_{\text{ensemble}} \) is computed.

\subsection{Robustness to Adaptive Attacks}
\label{Robustness}

Following best practices established by Athalye et al.~\cite{ObfuscatedGG}, we evaluate \texttt{MirrorCheck} under strong white-box adaptive attacks to ensure our defense does not rely on gradient obfuscation or hidden randomness. We design a worst-case attacker with full knowledge of the victim model $\mathcal{F}_\theta$, the entire model zoo of image encoders $\{\mathcal{I}_{\phi_j}\}_{j=1}^N$, the text-to-image (T2I) models $\{G_\psi\}$, and the detection mechanism. 

\noindent The attacker’s goal is twofold: (i) to induce the victim model to produce a target text or class, and (ii) to evade detection by maximizing the similarity between the adversarial image and its regenerated counterpart. A practical challenge is that the victim's VLM output is inherently discrete; it produces sampled tokens, breaking the gradient path. To resolve this, we extend the classic Backward Pass Differentiable Approximation (BPDA)~\cite{ObfuscatedGG} method: the attacker uses the \textbf{true} discrete text in the forward pass to condition the T2I model $G_\psi$, but replaces the non-differentiable text sampling with a learnable Adapter $\mathcal{A}$ in the backward pass. This Adapter maps the VLM’s continuous hidden state to the T2I conditioning space, providing a differentiable surrogate path for gradients. Additionally, we apply Expectation over Transformation (EoT) \cite{EOT} to handle all sources of randomness in the pipeline, averaging over the stochastic sampling noise $\eta$ in $G_\psi$, and the one-time-use (OTU) perturbations $\gamma$ applied to encoder weights. Because the attacker does not know which encoders will be sampled, they must optimize over the entire model zoo. The resulting adaptive attack objective is:

\begin{equation}
\min_{\delta: \|\delta\|_\infty \leq \varepsilon} \quad L_{\mathcal{F}_\theta}(x + \delta) + \lambda \cdot \mathcal{L}_{\text{det}},
\end{equation}
where the detection loss $\mathcal{L}_{\text{det}}$ is defined as:
\begin{align}
\mathcal{L}_{\text{det}} &= \mathbb{E}_{\psi,\, \eta,\, \gamma} \Big[ 1 - \frac{1}{N} \sum_{j=1}^{N} S(\mathbf{z}_1, \mathbf{z}_2) \Big],\\
\mathbf{z}_1 &= \mathcal{I}_{\phi_j + \gamma}(x+\delta) \nonumber, \\
\mathbf{z}_2 &= \mathcal{I}_{\phi_j + \gamma}(G_\psi(\mathcal{F}_\theta(x+\delta);\, \eta)) \nonumber.
\end{align}
In this formulation, $L_{\mathcal{F}_\theta}(x+\delta)$ denotes the attacker’s primary task objective that forces the victim VLM to output a desired target caption or class. For example, in the original attack setting~\cite{zhao2023evaluate}, this is defined by aligning the adversarial input with a target image generated from a surrogate image encoder $\mathcal{E}$:
$$
\argmin_{\delta:\|\delta\|_\infty \leq \varepsilon} 
\, d\big(\mathcal{E}(x+\delta),\, \mathcal{E}(x_{\text{ref}})\big),
$$
where \text{$x_\text{ref}$} is the target image.The attacker minimizes the expected detection score over all relevant randomness to robustly evade detection in the worst case. To train the Adapter \( \mathcal{A} \), we construct a dataset of 34{,}000 paired feature representations from clean ImageNet images. \vspace{0.5em}

\noindent \textbf{Image-to-text (VLM) features.} For each image, we use UniDiffuser as a captioning pipeline to generate a caption and extract the corresponding intermediate captioning representation by sampling from a trained DPM solver and encoding it via the model’s decoder head, yielding \( \mathbf{z}_{\text{VLM}} \in \mathbb{R}^{77 \times 64} \). \vspace{0.5em}

\noindent \textbf{Text-to-image (T2I) features.} The generated caption is then passed to a Stable Diffusion pipeline, from which we extract the internal prompt embedding \( \mathbf{z}_{\text{T2I}} \in \mathbb{R}^{77 \times 768} \), using deterministic settings (1 denoising step). The Adapter \( \mathcal{A} \) is implemented as a lightweight convolutional MLP: three convolutional layers process the input \( \mathbf{z}_{\text{VLM}} \) as a 2D tensor of shape \( 1 \times 77 \times 64 \), with ReLU activations and max-pooling. The resulting feature map is flattened and passed through four fully connected layers to produce a final output of shape \( 77 \times 768 \), matching the T2I embedding dimensionality. We optimize the Adapter using an L2 regression loss:
\[
\mathcal{L}_{\text{Adapter}} = \left\| \mathcal{A}(\mathbf{z}_{\text{VLM}}) - \mathbf{z}_{\text{T2I}} \right\|^2_2.
\]

\begin{table*}[t]
\caption{\small{\textbf{Adversarial detection performance across VLM attacks.} Our MirrorCheck variants consistently outperform both unimodal approaches and multimodal VLM-specific methods, achieving superior detection rates as high as 0.99. Best results in \textbf{bold}.}}
\vspace{-1.5em}
\label{tab:method_comparison}
\begin{center}
     \setlength\tabcolsep{6pt}
\resizebox{\linewidth}{!}{
\begin{tabular}{llcccccccccccc} 
\toprule
\multirow{2}{*}{Victim Model} & \multirow{2}{*}{Attack Setting} & \multicolumn{4}{c}{Unimodal Approaches} & \multicolumn{6}{c}{Multimodal Approaches} & \multicolumn{2}{c}{Ours} \\
\cmidrule(lr){3-6} \cmidrule(lr){7-12} \cmidrule(lr){13-14}
& & FS & MagNet & PuVAE & DiffPure & CIDER & Naive & CLIP & JailGuard & SmoothVLM & DPS & MC & Stochastic-MC \\ 
\midrule
\multirow{2}{*}{UniDiffuser} & AttackVLM-T & 0.56 & 0.74 & 0.51 & 0.80 & 0.84 & 0.68 & 0.59 & 0.81 & 0.82 & 0.83 & \textbf{0.96} & 0.95 \\ 
\cdashline{2-14}
& AttackVLM-Q& 0.65 & 0.85 & 0.70 & 0.81 & 0.80 & 0.65 & 0.57 & 0.83 & 0.83 & 0.85 & \textbf{0.98} & \textbf{0.98} \\ 
\midrule
\multirow{2}{*}{BLIP} & AttackVLM-T & 0.52 & 0.60 & 0.50 & 0.71 & 0.81 & 0.66 & 0.61 & 0.79 & 0.77 & 0.81 & 0.90 & \textbf{0.93} \\ 
\cdashline{2-14}
& AttackVLM-Q& 0.57 & 0.65 & 0.80 & 0.76 & 0.85 & 0.64 & 0.55 & 0.84 & 0.81 & 0.84 & 0.89 & \textbf{0.97} \\ 
\midrule
\multirow{3}{*}{BLIP-2} & AttackVLM-T & 0.61 & 0.73 & 0.52 & 0.80 & 0.84 & 0.70 & 0.62 & 0.82 & 0.80 & 0.86 & 0.93 & \textbf{0.94} \\ 
\cdashline{2-14}
& AttackVLM-Q& 0.61 & 0.85 & 0.72 & 0.83 & 0.77 & 0.67 & 0.58 & 0.80 & 0.78 & 0.83 & 0.92 & \textbf{0.99} \\
\cdashline{2-14}
& Attack-Bard & - & - & - & 0.79 & 0.87 & 0.65 & 0.58 & 0.89 & 0.87 & 0.95 & \textbf{0.98} & 0.95 \\
\midrule
\multirow{2}{*}{Img2Prompt} & AttackVLM-T & 0.51 & 0.56 & 0.50 & 0.67 & 0.83 & 0.61 & 0.56 & 0.83 & 0.83 & 0.86 & 0.79 & \textbf{0.90} \\ 
\cdashline{2-14}
& AttackVLM-Q& - & 0.65 & 0.78 & 0.69 & 0.79 & 0.60 & 0.55 & 0.81 & 0.74 & 0.82 & 0.85 & \textbf{0.92} \\ 
\midrule
LLaVA & Attack-MMFM & - & - & - & 0.67 & 0.83 & 0.62 & 0.52 & \textbf{0.85} & \textbf{0.85} & \textbf{0.85} & 0.82 & \textbf{0.85} \\
\midrule
OpenFlamingo & Attack-MMFM & - & - & - & 0.65 & 0.84 & 0.60 & 0.51 & \textbf{0.87} & 0.84 & 0.86 & 0.81 & 0.81 \\
\midrule
MiniGPT-4 & AttackVLM-T & 0.54 & 0.51 & 0.53 & 0.62 & \textbf{0.85} & 0.57 & 0.51 & \textbf{0.85} & 0.80 & \textbf{0.85} & 0.66 & 0.67 \\
\bottomrule
\end{tabular}
}
\end{center}
\end{table*}

\section{Experiments}
\label{exp}

We evaluate \texttt{MirrorCheck} variants across three key dimensions: (1) performance in unimodal and multimodal tasks, (2) comparison against baselines, and (3) robustness against adaptive attacks. All experiments are run three times on 2000 images (1000 clean and 1000 attacked) and use open-source models for reproducibility.

\subsection{Implementation Details}
\label{sec:implementation}

\paragraph{Victim Models.} We evaluate on diverse architectures: \textit{Multimodal models} including UniDiffuser \cite{bao2023one}, BLIP \cite{li2022blip}, Img2Prompt \cite{10204235}, BLIP-2 \cite{li2023blip}, LLaVA \cite{liu2023visual}, OpenFlamingo \cite{awadalla2023openflamingo}, and MiniGPT-4 \cite{zhu2023minigpt}; and \textit{Unimodal models} including DenseNet \cite{densenet} and MobileNet \cite{mobilenetv2} for classification tasks.
\vspace{-1em}
\paragraph{Adversarial Attack Settings.} We evaluate against various attack strategies using settings from their original papers: \textit{Targeted attacks} such as AttackVLM transfer and query-based variants \cite{zhao2023evaluate} on ImageNet-1K validation images \cite{deng2009imagenet} with randomly selected MS-COCO captions \cite{lin2014microsoft} as targets; \textit{Untargeted attacks} including Attack-Bard \cite{bard} on NIPS17 dataset \cite{nips17}, Attack-MMFM \cite{untarget} on COCO 2014 captioning tasks \cite{lin2014microsoft}, and standard attacks (FGSM \cite{goodfellow2015explaining}, BIM \cite{kurakin2018adversarial}, PGD \cite{madry2019deep}, DeepFool \cite{deepfool}, C\&W) on CIFAR-10 \cite{cifar10} and ImageNet \cite{deng2009imagenet}. All attack parameters follow the original implementations.
\vspace{-1em}
\paragraph{T2I Models.} Our T2I model zoo includes Stable Diffusion v1.4/v1.5 \cite{rombach2022highresolution}, UniDiffuser \cite{bao2023one}, and ControlNet \cite{zhang2023adding}. All generation uses 50 timesteps, producing $512 \times 512$ pixel outputs. For \texttt{Stochastic MirrorCheck}, one T2I model is randomly selected per inference.
\vspace{-1em}
\paragraph{Image Encoders.} Our encoder collection includes OpenAI CLIP \cite{clip} variants, OpenCLIP \cite{openclip} models, and pretrained VGG16 \cite{vgg} and ResNet-50 \cite{resnet} from PyTorch. Model-specific preprocessing is applied to both input and generated images. For \texttt{Stochastic MirrorCheck}, $n \in \{1,3,5,7,10\}$ encoders are randomly selected per inference.
\vspace{-1em}
\paragraph{MirrorCheck Config.} \texttt{Vanilla MirrorCheck} employs fixed T2I models (primarily SD v1.4) and predetermined encoder sets. \texttt{Stochastic MirrorCheck} introduces randomization across three dimensions: (1) random T2I model selection from our zoo, (2) random encoder selection (single or ensemble), and (3) One-Time-Use (OTU) perturbations with Gaussian noise scales $\{5e^{-6}, 5e^{-4}, 1e^{-3}\}$ applied to encoder parameters before each inference.
\vspace{-1em}
\paragraph{Baseline.} We evaluate against both classical detection approaches adapted from image classification and recent VLM-specific defense methods. (\textbf{Unimodal Baselines.}) \textbf{MagNet} \cite{magnet} uses autoencoders to detect adversarial inputs via reconstruction error analysis on the manifold of clean images. \textbf{PuVAE} \cite{pu2016variational} employs variational autoencoders for adversarial purification through latent space projection. \textbf{FeatureSqueeze} \cite{FeatureSqueeze} creates "squeezed" input versions and compares model predictions for discrepancy detection. \textbf{DiffPure} \cite{DiffPure} leverages diffusion models for adversarial purification. For VLM adaptation, we train MagNet and PuVAE autoencoders on ImageNet to learn clean image manifolds, while FeatureSqueeze compares captions generated from original versus squeezed inputs. (\textbf{Multimodal Baselines.}) \textbf{CIDER} \cite{xu2024cider} detects adversarial examples in Multimodal Large Language Models by measuring the discrepancy in cross-modal semantic alignment between text-image pairs before and after diffusion-based denoising. \textbf{Naive} implements an ensemble voting approach where the same input is passed to five VLMs (including the victim model), flagging inputs with significant response divergence as adversarial. \textbf{CLIP} calculates similarity scores between input images and victim model responses using CLIP embeddings \cite{clip} for detection. \textbf{JailGuard} \cite{zhang2023jailguard}, \textbf{SmoothVLM} \cite{sun2024safeguarding}, and \textbf{DPS} \cite{zhou2024defending} are mutation-based approaches defending against jailbroken prompts. We adapt JailGuard and SmoothVLM by mutating inputs using filters and patch-based perturbations, while DPS leverages cropping techniques. All methods flag inputs based on response divergence after mutation.

\subsection{Benchmarking \texttt{MirrorCheck}'s Performance}
\Cref{SD-similarity} shows similarity scores using \texttt{Stochastic MirrorCheck} across diverse victim models and attack scenarios with randomized encoder selection and \texttt{OTU} perturbations. Clean images consistently achieve high similarity scores, demonstrating that legitimate inputs maintain strong semantic consistency when processed through our stochastic detection framework. In contrast, adversarial examples exhibit lower similarity scores. This gap between clean and adversarial similarity scores enables effective detection. The results demonstrate robust performance across various ensemble sizes (1, 3, 5, 7, 10 encoders) and noise perturbation scales (5e-6, 5e-4, 1e-3). Notably, larger ensemble sizes generally provide more stable detection boundaries, while different noise scales offer varying levels of adaptive attack resistance. Leveraging the observed similarity scores, we compute the \emph{detection accuracy} of our method as the ratio of correctly identified clean and adversarial images to the total number of images for each attack. As shown in ~\Cref{SD-detection}, the method maintains consistent discriminative capabilities and detection accuracies (as high as $99\%$) across unimodal and multimodal architectures under diverse attack settings and across different downstream tasks. We present complete similarity scores and detection results using different encoders and T2I models for our vanilla variant in \Cref{abla_other}.
 % \textcolor{red}{CREF}.

\subsection{Comparison with Baselines}
\Cref{tab:method_comparison} presents adversarial detection performance across a diverse set of vision-language models (VLMs) and attack scenarios. We compare our proposed MirrorCheck variants (\textbf{MC} and \textbf{Stochastic-MC}) against both unimodal defenses (FS, MagNet, PuVAE, DiffPure) and multimodal methods specifically tailored for VLMs (CIDER, Naive, CLIP, JailGuard, SmoothVLM, DPS). Overall, MirrorCheck consistently achieves superior detection rates, often by a large margin. For example, on UniDiffuser under the AttackVLM-Q setting, Stochastic-MC attains a detection score of \textbf{0.98}, outperforming the next best baseline by more than 0.13. Similarly, on BLIP-2 with AttackVLM-Q, MirrorCheck reaches \textbf{0.99}. Our approach demonstrates robustness across both text- and query-based attacks, as well as multimodal fusion attacks (e.g., Attack-MMFM), where existing unimodal defenses fail to generalize. Notably, even against strong baselines like DiffPure and JailGuard, which are widely used in current multimodal adversarial defenses, MirrorCheck maintains consistent gains without requiring model retraining or access to internal gradients/logits. This highlights the practicality and scalability of our method for real-world deployment.

\subsection{Impact of Scaling Factor}
From \cref{SD-detection}, we find that using moderate amounts of noise (scaling factors $\leq 5\times10^{-4}$) works best for detecting attacks across all models and attack types. This happens because larger scaling factors disrupts the model's learned patterns, making it harder to distinguish between normal and malicious inputs. The same pattern holds for different types of attacks, though the effect is less noticeable for simpler classification attacks.

\subsection{Impact of Encoder Size}
We also observe that using more encoders generally improves attack detection rates. However, the benefits start to plateau after 5–7 encoders (this is not the case for adaptive attacks, see \cref{adapa}). This suggests that while having multiple encoders helps by providing diverse perspectives, adding too many encoders provides minimal improvement while increasing computational costs.

\subsection{Adaptive Attack} \label{adapa}
\begin{figure}[t]
    \centering
    \begin{subfigure}[b]{0.10\textwidth}
        \includegraphics[width=\textwidth]{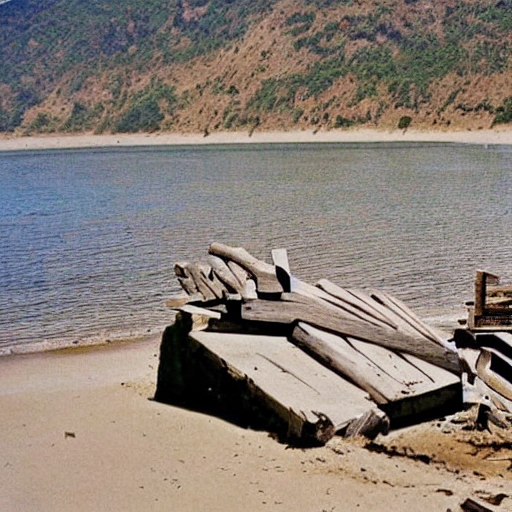}
    \end{subfigure}
    \hfill
    \begin{subfigure}[b]{0.10\textwidth}
        \includegraphics[width=\textwidth]{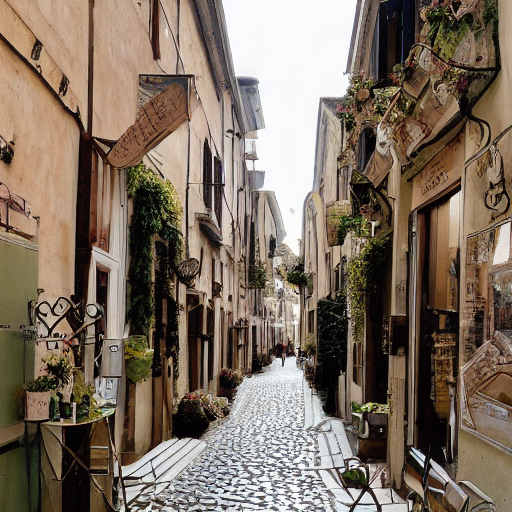}
    \end{subfigure}
    \hfill
    \begin{subfigure}[b]{0.10\textwidth}
        \includegraphics[width=\textwidth]{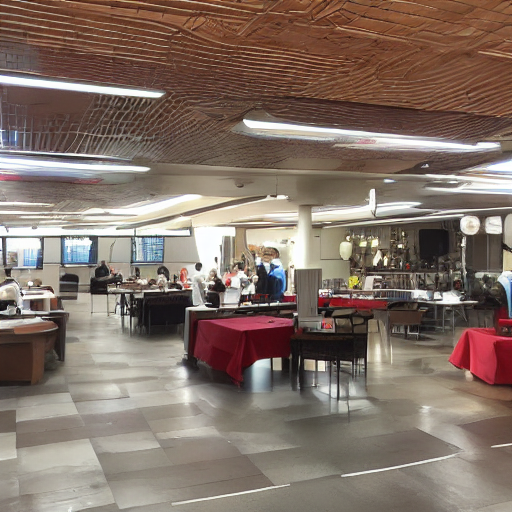}
    \end{subfigure}
    \hfill
    \begin{subfigure}[b]{0.10\textwidth}
        \includegraphics[width=\textwidth]{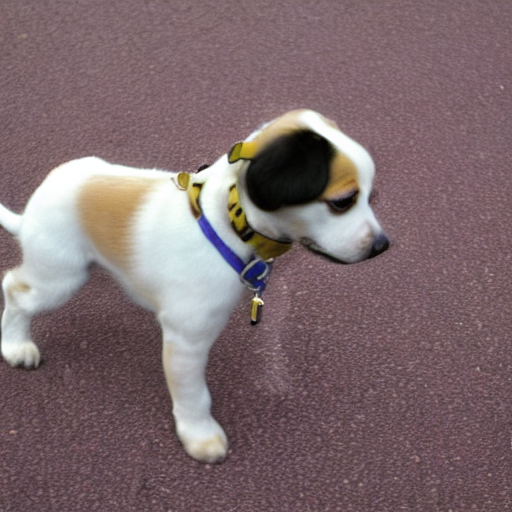}
    \end{subfigure}
    \caption{Images generated from Stable Diffusion conditioned on Adapter outputs computed from UniDiffuser’s captioning features.}
    \label{fig:adapter_samples}
\end{figure}

The Adapter \( \mathcal{A} \) is trained using the Adam optimizer over 500 epochs with a batch size of 64. To qualitatively verify the effectiveness of the trained Adapter, we generate images using Stable Diffusion by conditioning directly on the Adapter output \( \mathcal{A}(\mathbf{z}_{\text{VLM}}) \), where \( \mathbf{z}_{\text{VLM}} \) is the feature produced by UniDiffuser’s image-to-text captioning pipeline using ImageNet test images. As shown in \cref{fig:adapter_samples}, the resulting images are visually coherent and reflect the semantics of the original input images, demonstrating that the Adapter successfully bridges the representation gap between the captioning and generation pipelines.

\begin{figure}[t]
    \centering
\includegraphics[width=\linewidth]{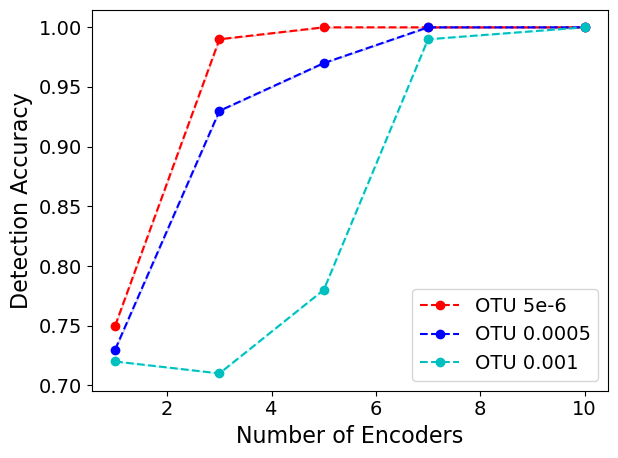}
    \caption{Average detection accuracy over three runs with different random seeds as a function of the number of encoders for different noise levels under adaptive attack.}
\label{fig:adaptive_accuracy_avg}
\end{figure}

\vspace{-1em}
\paragraph{Detection Accuracy Under Adaptive Attacks.} We evaluate our method under a white-box adaptive attack following the BPDA+EoT strategy described in \S\ref{Robustness}. We randomly sample 10 image–target pairs from ImageNet and optimize an \( \ell_\infty \)-bounded perturbation (\( \varepsilon = 8 \)) over 100 PGD steps with a step size of 1. The attacker has full access to the victim model, the trained Adapter, and the entire detection pipeline. The attack is implemented using Stable Diffusion (v1.5) for generation, UniDiffuser for captioning, and the model zoo of 10 CLIP and OpenCLIP encoders. We apply Stochastic MirrorCheck for detection. Figure~\ref{fig:adaptive_accuracy_avg} illustrates the accuracy trends. We observe that detection accuracy consistently improves with the number of encoders in the ensemble. With just 3 encoders, the accuracy already approaches 95\% for low and medium OTU noise (5e-6, 5e-4), and with 7 or more encoders, all configurations exceed 98\%. At the highest noise level (0.001), the improvement is more gradual, with accuracy increasing from ~71\% (single encoder) to ~99\% (10 encoders), highlighting the compounding benefits of both encoder diversity and stochasticity. These results confirm that even under strong adaptive attacks, increasing ensemble size and applying per-run perturbation noise significantly enhances detection robustness.

\subsection{Additional ablation and Insights}
We perform additional ablation studies and analyses to further explain why \ours{} is effective (\cref{sec:theory,abla_other}). We reinterpret our framework from an autoencoder perspective (\cref{AutoencoderView}) and highlight key intuitive observations (\cref{obs}) that reveal its strengths and potential failure cases. Our results show that \ours{} is model-agnostic and generalizes across diverse image encoders and T2I models. We also introduce an adaptive attack with a different objective (\cref{adaptappendix}) and show that \ours{} remains robust. Finally, we study the impact of the clean-to-adversarial ratio and provide qualitative visualizations showing \ours{} in action.

\subsection{Computational Efficiency}

Our experiments were carried out on a NVIDIA Quadro RTX A6000 48GB GPU. The entire defense pipeline takes approximately 15 seconds per image. Within this process, obtaining a caption from the victim VLM model takes around 0.2 seconds, generating an image takes about 5 seconds, and calculating similarity requires approximately 10 seconds. However, this is the worst case scenario and there are multiple methods to improve this time i.e., reducing timesteps for generation from 50 to 10 allows the pipeline process an image in just 1.2 seconds with a little compromise in detection performance.

% \section{Discussion} 
% \label{sec:discussion}

\section{Conclusion}
\label{sec:conclusion}

We introduce \texttt{MirrorCheck}, a novel adversarial detection framework that leverages T2I generation and similarity analysis. Our \texttt{Stochastic MirrorCheck} variant employs randomized model selection and One-Time-Use perturbations to create robust defenses against adaptive attacks. Comprehensive evaluation across diverse VLM architectures and attack scenarios demonstrates superior performance compared to adapted baseline methods, with detection accuracies going as high as 99\%. This work establishes a new paradigm for multimodal adversarial defense and provides a foundation for securing next-generation AI systems against evolving threats.

{
    \small
    \bibliographystyle{ieeetr}
    \bibliography{main}
}
% \clearpage 


\clearpage
\setcounter{page}{1}
\maketitlesupplementary

\appendix

\section{Further Background and Related Works}
\label{lrbck}
\subsection{Visual-Language Models (VLMs)}
Humans possess the remarkable ability to seamlessly integrate information from various sources concurrently. For instance, in conversations, we adeptly interpret verbal cues, body language, facial expressions, and intonation. Similarly, VLMs demonstrate proficiency in processing such multimodal signals, allowing machines to comprehend and generate image-related content that seamlessly merges visual and textual components. Contemporary VLM architectures such as CLIP \cite{clip} predominantly leverage transformer-based models \cite{vaswani2023attention, dosovitskiy2021image} for processing both images and text due to their effectiveness in capturing long-range dependencies. At the heart of the transformers lies the multi-head attention mechanism, which plays a pivotal role in these models' functionality.\vspace{0.5em}

\noindent To enable multimodal comprehension, VLMs typically comprise three key components: \textit{(i)} an Image Model responsible for extracting meaningful visual features from visual data, \textit{(ii)}  a Text Model designed to process natural language, and \textit{(iii)} a Fusion Mechanism to integrate representations from both modalities. Encoders in VLMs can be categorized based on their fusion mechanisms into Fusion encoders \cite{li2020oscar, li2021align, li2019visualbert, su2019vl}, which directly combine image and text embeddings, Dual encoders \cite{clip, li2022blip, li2023blip, jia2021scaling}, which process modalities separately before interaction, and Hybrid methods \cite{Singh2021FLAVAAF, bao2022vlmo} that leverage both approaches. Furthermore, fusion schemes for cross-modal interaction can be classified into single-stream \cite{li2020oscar, li2019visualbert, su2019vl, bao2022vlmo, Singh2021FLAVAAF} and dual-stream \cite{li2021align} architectures. The recent surge in multimodal development, driven by advances in vision-language pretraining (VLP) methods, has led to diverse vision-language applications falling into three main categories: \textit{(i)} Image-text tasks (such as image captioning, retrieval, and visual question answering), \textit{(ii)} Core computer vision tasks (including image classification, object detection, and image segmentation), and \textit{(iii)} Video-text tasks (such as video captioning, video-text retrieval, and video question-answering). 

\subsection{Other Adversarial Attacks used against VLMs}

\paragraph{Attack-Bard \citep{bard}.} For a victim model that is a Multimodal Large Language Model (MLLM), adversarial examples that effectively perturb the image embeddings of Bard \citep{gemini} will consequently impact the text generation process. Let \( x\) represent a natural image and \( \tilde{\mathcal{I}}_{i\phi}()\) be a set of surrogate image encoders. The image embedding attack is defined as solving the following optimization problem:

\begin{equation} 
\label{bard}
\text{argmax}_{\delta: \|\delta\|_{\infty} \leq \varepsilon }   \sum_{i=1}^N \|\tilde{\mathcal{I}}_{i\phi}(x_{adv}) -\tilde{\mathcal{I}}_{i\phi(x)}\|_2^2
\end{equation}
where $x_{adv} = x + \delta$ and the goal is to maximize the difference between the embeddings of the adversarial image \( x_{adv} \) and the natural image \( x \) while ensuring that the perturbation \( \delta \) remains within a specified threshold \( \epsilon \). To address the optimization problem in \eqref{bard}, \cite{bard} employed the SSA-CWA approach, as introduced in \cite{chen2023rethinking}.

\vspace{-1em}
\paragraph{Attack-MMFM \citep{untarget}.} An untargeted attack proposed against multimodal foundation models. To introduce minor perturbations to the visual inputs of a VLM, the authors propose a white-box untargeted attack. Specifically, given a natural image $x$, a ground truth caption $t$, along with context images $c$ and context text $z$, the objective is to design an attack that increases the negative log-likelihood of the target text $t^*$ within the constraints of the threat model:

\begin{equation}
\label{robustvlm}
    \max_{\delta_x, \delta_c} - \sum_{i=1}^{m} \log p(t^*_i \mid t^*_{<i}, z, x + \delta_x, c + \delta_c)
\end{equation}
\[
\text{s.t. } \|\delta_x\|_{\infty} \leq \epsilon_x, \|\delta_c\|_{\infty} \leq \epsilon_c
\]
In equation \ref{robustvlm} above, $\delta_x$ is the perturbation to the input image and $\delta_c$ is the perturbation to the context images. In the setting where only the input images are attacked, optimization is performed only on $\delta_x$ and $\epsilon_c = 0$.

\subsection{Adversarial Attacks used for Classification}
\label{related-a}
An adversarial example,  within the scope of machine learning, is a sample intentionally manipulated by an adversary to provoke an incorrect output from a target classifier. Typically, in image classification tasks, where the ground truth is based on human perception, defining adversarial examples involves perturbing a correctly classified sample (referred to as the seed example) by a limited amount to generate a misclassified sample (denoted as $x_{\text{adv}}$). Existing research on adversarial example generation predominantly centers on image classification models, reflecting the prominence and vulnerability of such models to adversarial attacks. Numerous methodologies have been introduced to craft adversarial examples, encompassing fast gradient-based techniques \cite{goodfellow2015explaining,liu2016delving}, optimization-based strategies \cite{szegedy2013intriguing,carlini2017towards}, and other innovative approaches \cite{nguyen2015deep,papernot2016limitations}. Notably, \cite{carlini2017towards} introduced state-of-the-art attacks that impose constraints on L$_0$, L$_2$, and L$_{\infty}$ norms, highlighting the versatility and effectiveness of adversarial attacks across various norm spaces.\vspace{0.5em}

\noindent Adversarial examples can be categorized as targeted or untargeted depending on the adversary's objective. In targeted attacks, the adversary aims for the perturbed sample $x_{\text{adv}}$ to be classified as a specific class, while in untargeted attacks, the objective is for $x_{\text{adv}}$ to be classified as any class other than its correct class. \vspace{0.5em}

\noindent Formally, a targeted adversary seeks to find an $x_{\text{adv}}$ such that the target classifier assigns it to the target class $y$ while remaining within a certain distance $\epsilon$ from the original sample $x_{\text{clean}}$. Conversely, an untargeted adversary aims to find an $x_{\text{adv}}$ which is misclassified compared to the original $x_{\text{clean}}$ within the same distance threshold $\epsilon$. The adversary's strength, denoted as $\epsilon$, restricts the allowable transformations applied to the seed example. In contrast, the distance metric $\Delta(x_{\text{clean}},x_{\text{adv}})$ and the threshold $\epsilon$ model how close an adversarial example needs to be to the original to deceive a human observer. As specified in our related work section, we will introduce some attack strategies used in classification tasks. We also leverage these attacks to test the efficacy of \ours{} in this setting;
\begin{itemize}
    \item \textbf{Fast Gradient Sign Method} (FGSM, $L_\infty$, Untargeted): The Fast Gradient Sign Method (FGSM) is an adversarial attack technique proposed by Goodfellow et al. \cite{goodfellow2015explaining} that efficiently generates adversarial examples for deep neural networks (DNNs). The objective of the FGSM attack is to perturb input data in such a way that it induces misclassification by the target model while ensuring the perturbations are imperceptible to human observers. The main idea behind FGSM is to compute the gradient of the loss function with respect to the input data, and then perturb the input data in the direction that maximizes the loss. Specifically, FGSM calculates the gradient of the loss function with respect to the input data, and then scales the gradient by a small constant $\epsilon$ to determine the perturbation direction. This perturbation is added to the original input data to create the adversarial example. Mathematically, the FGSM perturbation is defined as:
    \[
        x_{\text{adv}} = x_{\text{clean}} + \epsilon \cdot \text{sign}(\nabla_x J(w^Tx_{\text{clean}}), y))
    \]
    where $\epsilon$ is a small constant controlling the magnitude of the perturbation, and sign denotes the sign function. The objective function of the FGSM attack is typically the cross-entropy loss between the predicted and true labels, as it aims to maximize the model's prediction error for the given input.

    \item \textbf{Basic Iterative Method} (BIM, $L_\infty$, Untargeted): The Basic Iterative Method (BIM) attack \cite{feinman2017detecting}, also known as the Iterative Fast Gradient Sign Method (IFGSM), is an iterative variant of the FGSM attack designed to generate stronger adversarial examples. Like FGSM, the objective of the BIM attack is to craft adversarial perturbations that lead to misclassification by the target model while remaining imperceptible to human observers. In the BIM attack, instead of generating a single perturbation in one step, multiple small perturbations are iteratively applied to the input data. This iterative approach allows for finer control over the perturbation process, resulting in adversarial examples that are more effective and harder for the target model to defend against. The BIM attack starts with the original input data and applies small perturbations in the direction of the gradient of the loss function with respect to the input data. After each iteration, the perturbed input data is clipped to ensure it remains within a small $\epsilon$-ball around the original input. This process is repeated for a fixed number of iterations or until a stopping criterion is met. Mathematically, the perturbed input at each iteration $s$ of the BIM attack is given by:

    \[
        x_{\text{adv}}^{s} = \text{clip}_{\epsilon}(x_{\text{adv}}^{s-1} + \alpha \cdot \text{sign}(\nabla_{x} J(w^Tx_{\text{clean}}), y))
    \]

    where $\text{Clip}_{\epsilon}$ denotes element-wise clipping to ensure the perturbation magnitude does not exceed $\epsilon$, and $\alpha$ is a small step size controlling the magnitude of each perturbation. The BIM attack aims to maximize the loss function while ensuring the perturbations remain bounded within the $\epsilon$-ball around the original input.

    \item \textbf{DeepFool} ($L_2$, Untargeted): The DeepFool attack \cite{deepfool} is an iterative and computationally efficient method for crafting adversarial examples. It operates by iteratively perturbing an input image in a direction that minimally changes the model's prediction. The objective of the DeepFool attack is to find the smallest perturbation that causes a misclassification while ensuring that the adversarial example remains close to the original input in terms of the $L_2$-norm. The DeepFool attack starts with the original input image and iteratively computes the perturbation required to push the image across the decision boundary of the model. It computes the gradient of the decision function with respect to the input and then finds the direction in which the decision boundary moves the most. By iteratively applying small perturbations in this direction, the DeepFool attack gradually moves the input image towards the decision boundary until it crosses it. Mathematically, the perturbed input at each iteration of the DeepFool attack is computed as follows:

    \[
        x_{\text{adv}}^{s} = x_{\text{adv}}^{s-1} + \alpha \cdot \frac{\nabla_{\text{f}}(x_{\text{clean}})}{\|\nabla_{\text{f}}(x_{\text{clean}})\|}_2
    \]

    where $ x_{\text{adv}}^{s-1}$ is the input image at the current iteration $s$, \(\alpha\) is a small step size, and \(\nabla_{\text{f}}(x)\) is the gradient of the decision function with respect to the input image \(x_{\text{clean}}\). The process continues until the model misclassifies the perturbed input or until a maximum number of iterations is reached.

    \item \textbf{Projected Gradient Descent} (PGD, $L_{2}$, Untargeted): The Projected Gradient Descent (PGD) attack \cite{madry2018towards} is an advanced iterative method used for crafting adversarial examples. It builds upon the Basic Iterative Method (BIM), extending it by continuing the perturbation process until reaching a specified maximum perturbation magnitude. The objective of the PGD attack is to find the smallest perturbation that leads to misclassification while constraining the perturbed example to remain within a specified $L_p$-norm distance from the original input. The PGD attack starts with the original input image and iteratively computes the perturbation required to induce misclassification. At each iteration, it calculates the gradient of the loss function with respect to the input and applies a small step in the direction that maximizes the loss while ensuring the perturbed example remains within the specified $L_p$-norm ball around the original input. This process continues for a predetermined number of iterations or until a misclassification is achieved. Mathematically, the perturbed input at each iteration of the PGD attack is computed as follows:

    \begin{align*}
x_{\text{adv}}^{s} &=
\operatorname{clip}\!\Bigl(
x_{\text{adv}}^{s-1}
+ \alpha\,\mathrm{sign}\!\bigl(\nabla_{x} J(w^{\top} x_{\text{clean}}, y)\bigr),\\
&\quad x_{\text{adv}}-\epsilon,\,
x_{\text{adv}}+\epsilon
\Bigr)
\end{align*}

    where $x_{\text{adv}}^{t-1}$ is the input image at the current iteration $t$, \(\alpha\) is the step size, $\nabla_{\text{x}} J(w^Tx_{\text{clean}}, y)$ is the gradient of the loss function with respect to the input image $x_{\text{clean}}$, and \(\text{clip}\) function ensures that the perturbed image remains within a specified range defined by the lower and upper bounds.

    \item \textbf{Carlini-Wagner} (C\&W, $L_{2}$, Untargeted): The Carlini-Wagner (C\&W) attack \cite{carlini2017towards}, introduced by Carlini and Wagner in 2017, is a powerful optimization-based method for crafting adversarial examples. Unlike many other attack methods that focus on adding imperceptible perturbations to input data, the C\&W attack formulates the attack as an optimization problem aimed at finding the smallest perturbation that leads to misclassification while satisfying certain constraints. The objective of the C\&W attack is to find a perturbation $\delta$ that minimizes a combination of the perturbation magnitude and a loss function, subject to various constraints. The loss function is typically designed to encourage misclassification while penalizing large perturbations. The constraints ensure that the perturbed example remains within a specified $L_p$-norm distance from the original input and maintains perceptual similarity. The objective function of the C\&W attack can be formulated as follows:
    
    \[
        \min \|\delta\|_l + c \cdot f(x_{\text{clean}} + \delta)
    \]
    
    where $\|\delta\|_l$ represents the $L_l$-norm of the perturbation, $f(x_{\text{clean}} + \delta)$ is the loss function representing misclassification, and $c$ is a regularization parameter that balances the trade-off between the perturbation magnitude and the loss function.
\end{itemize}

\section{Analysis}
\label{sec:theory}

In this section, we provide an explanation for why our proposed \ours{} pipeline is robust to both \emph{non-adaptive} and \emph{adaptive} adversarial attacks. We begin by propose \ours{} as an autoencoder before defining the overall setting, threat model, and key assumptions. We then present our main theorem and proof sketch, demonstrating why clean images remain undetected while adversarial ones are flagged with high probability. Finally, we extend the argument to the \emph{adaptive} setting, where the attacker has knowledge of our detection pipeline and attempts to circumvent it by making the entire process differentiable.

\subsection{\ours{} as an Autoencoder}
\label{AutoencoderView}
In the auto-encoder literature, reconstruction error has been shown to be a reliable indicator of whether a sample is in or out of the training distribution~\cite{zhou2022rethinking, durasov2023zigzag,durasov2024enabling}.  We now cast  \ours{} as a particular kind of auto-encoder to leverage these results and justify our approach. \ours{} can be conceptualized within the structure of regular~\cite{hinton2006reducing, vincent2010stacked, makhzani2016} and Variational Autoencoders (VAEs)~\citep{Kingma2014, burda2015importance, higgins2017beta}, which typically encode input data into a continuous latent space through an encoder and reconstruct the input using a decoder. Unlike typical variational-autoencoders, \ours{} relies on a discrete, categorical latent space comprising textual descriptions generated from images. In this respect, it is in line with recent VAEs that incorporate categorical latent variables through mechanisms such as the Gumbel-Softmax distribution~\cite{maddison2016concrete, jang2017categorical, baevski2020wav2vec, sadhu2021wav2vec,gangloff2022leveraging}. \\

\noindent The I2T phase of \ours{} acts as the encoder, mapping high-dimensional visual data into a discrete latent space represented by text. This process can be mathematically expressed as 
\begin{equation}
q_\phi(\mathbf{z} | \mathbf{x}) = \text{Cat}(\mathbf{z}; \boldsymbol{\pi}(\mathbf{x})) \; ,
\end{equation}
where $\mathbf{x}$ is the input image, $\mathbf{z}$ represents the latent textual description, $\text{Cat}$ denotes the categorical distribution, and $\boldsymbol{\pi}(\mathbf{x})$ is the distribution over the discrete latent variables conditioned on the input image, parameterized by $\phi$.

\noindent The T2I phase serves as the decoder. It reconstructs the visual data from these textual descriptions. It can be written as
\begin{equation}
p_\theta(\mathbf{x} | \mathbf{z}) = \text{Bernoulli}(\mathbf{x}; \boldsymbol{\sigma}(\mathbf{z})),
\end{equation}
where $\boldsymbol{\sigma}(\mathbf{z})$ models the probability of generating an image $\mathbf{x}$ from the latent description $\mathbf{z}$, parameterized by $\theta$. When sampling caption text with a non-zero softmax temperature, these steps resemble the Gumbel-Softmax reparameterization trick, typically used in Variational Autoencoders (VAEs) to sample from the latent~\cite{maddison2016concrete, jang2017categorical}.\\

\noindent Thus, using the reconstruction error as an indication of whether an input has been compromised via an adversarial attack is as justified as using it to determine if a sample is out-of-distribution when employing a VAE. This aligns with earlier work~\cite{magnet,pu2016variational} that showed that this metric is good at detecting adversarial attacks. It is also in the same spirit as approaches to detecting anomalies through segmentation and reconstruction ~\cite{Lis19, Lis24}. 

\subsection{Key Observations} \label{obs}
\begin{enumerate}
    \item \emph{Clean Consistency.}  
    A small perturbation from $x_{\text{clean}}$ typically does not change the VLM’s caption drastically. If the caption remains accurate, then $G_\psi$ produces a corresponding $x_{\text{gen,clean}}$ that is semantically aligned with $x_{\text{clean}}$. Finally, by Lipschitzness, these two images remain close in the embedding space, pushing $\mathrm{sim}(\cdot,\cdot)$ above $\tau^*$.

    \item \emph{Adversarial Detection.}  
    If the adversary’s perturbation $\delta$ drastically changes the VLM’s output caption $t_{\text{adv}}$, then $x_{\text{gen,adv}}$ becomes semantically inconsistent with $x_{\text{adv}}$. In embedding space, $\mathrm{sim}\bigl(\mathcal{I}_{\hat{\phi}}(x_{\text{adv}}),\,\mathcal{I}_{\hat{\phi}}(x_{\text{gen,adv}})\bigr)$ plummets below $\tau^*$, triggering detection.

    \item \emph{Randomization / Ensemble.}  
    Even if the adversary tries to \emph{adapt} by directly optimizing similarity, randomization or multiple encoders ensure that the gradient alignment is broken. The attacker cannot perfectly maintain high similarity under \emph{all} encoders, especially if one-time noise (OTU) is added right before inference.

    \item \emph{Ensemble Randomization Fractures Gradient Alignment.}  
    Different encoders $\{\mathcal{I}_{\phi j}\}$ exhibit distinct embedding geometries; a single perturbation $\delta$ usually cannot keep all similarity scores high at once. Unless $\|\delta\|$ is made tiny (reducing the adversarial effect), mismatch arises in at least one encoder, causing detection.

    \item \emph{OTU Noise Breaks Perfect Differentiability.}  
    If the defender applies random perturbations $\gamma$ to $\phi$ right before detection, the attacker’s precomputed gradient w.r.t.\ $\phi$ no longer matches the final inference pass. Thus, any carefully crafted $\delta$ may fail to preserve similarity under the \emph{real} detector.

    \item \emph{Semantic Mismatch Argument Persists.}  
    Even with a continuous pipeline, forcing a drastically different caption (to meet the adversarial goal) yields a T2I-generated image that is semantically far from $x_{\text{adv}}$. The attacker faces a contradiction between requiring \emph{large semantic drift} (to produce a malicious caption) and \emph{small semantic drift} (to retain high similarity). They typically cannot satisfy both simultaneously.
\end{enumerate}

% #######################################################################################################################################
\section{Additional Empirical Results}
\label{abla_other}
\paragraph{Generalization across Encoders and T2I Models.} \label{gen} We present the results of the vanilla variant of \ours{}. \cref{tab:sim} and \cref{tab:Detectionresults} shows the performance. Here, we leverage UniDiffuser T2I model \cite{bao2022all} and ControlNet \cite{zhang2023adding}. We observe better accuracies using UniDiffuser, compared to using Stable Diffusion. We also observe better accuracies using ControlNet, compared to using Stable Diffusion, and slightly better overall accuracies compared to UniDiffuser. \cref{unidiff-sim} and \cref{control-sim} show the similarities when using UniDiffuser-T2I \cite{bao2022all} and ControlNet \cite{zhang2023adding} for image generation and the CLIP models for evaluation, while \cref{tab:DetectionresultsUD} and \cref{tab:DetectionresultsCN} show the detection accuracies. Overall, we show that our approach is agnostic of and generalizes across encoders and T2I models.
%######################################################

% #######################################################################################################################################
% #######################################################################################################################################

\begin{table}[t]
\caption{\small{Similarity scores. The average shows that \ours{} is able to maximize the difference between clean and adversarial images for all victim models.}}
\label{tab:sim}
\begin{center}
\small
\resizebox{\linewidth}{!}{
\begin{tabular}{llccccccccccccccc}
\toprule
\multicolumn{1}{l}{\multirow{2}{*}{Victim Model}}   & \multicolumn{1}{l}{\multirow{2}{*}{Setting}} & \multicolumn{3}{c}{RN50}     & \multicolumn{3}{c}{RN101}    & \multicolumn{3}{c}{ViT-B/16} & \multicolumn{3}{c}{ViT-B/32} & \multicolumn{3}{c}{ViT-L/14}   \\ 
&    & Avg  & Min  & Max         & Avg  & Min  & Max            & Avg     & Min     & Max      & Avg     & Min     & Max      & Avg     & Min     & Max           \\ \midrule
\multirow{3}{*}{UniDiffuser} & Clean & \textbf{0.720} & 0.241 & 0.931 & \textbf{0.818} & 0.512 & 0.963 & \textbf{0.758}  & 0.320  & 0.975  & 0.750 & 0.344  & 0.973  & \textbf{0.723} & 0.244  & 0.952  \\\cdashline{3-17}
                            & AttackVLM-T    & 0.414  & 0.118 & 0.872  & 0.628 & 0.434  & 0.938 & 0.515 & 0.222 &0.852 & \textbf{0.807} & 0.426 & 0.958 & 0.516 & 0.130  & 0.820             \\\cdashline{3-17}
                             & AttackVLM-Q  & 0.421  & 0.165 & 0.742  & 0.676 & 0.539  & 0.780 & 0.551 & 0.330 & 0.759 & 0.528  & 0.274  & 0.725  &  0.547 & 0.280  & 0.735               \\ \midrule
\multirow{3}{*}{BLIP}        & Clean & \textbf{0.699} & 0.162 & 0.911 & \textbf{0.804} & 0.434 &0.953 & \textbf{0.741} & 0.247 & 0.948 & \textbf{0.723}  & 0.222 & 0.945 & \textbf{0.705} & 0.126 & 0.944\\\cdashline{3-17}
                            & AttackVLM-T      & 0.395 & 0.077 & 0.823   & 0.627 & 0.455 & 0.858   & 0.522 & 0.239 & 0.847    & 0.487 & 0.173 & 0.798   & 0.512 & 0.070 &  0.828        \\\cdashline{3-17}
                             & AttackVLM-Q     & 0.443 & 0.165 & 0.694 & 0.679  & 0.522  & 0.81 & 0.563 & 0.276  & 0.740 &  0.534 & 0.212  & 0.750 & 0.561 &  0.277 & 0.757              \\ \midrule
\multirow{3}{*}{BLIP-2}      & Clean  & \textbf{0.712}  & 0.151  & 0.936  & \textbf{0.813}  & 0.422  & 0.965 & \textbf{0.757} & 0.248 &0.961 & \textbf{0.737} & 0.213 & 0.946 & \textbf{0.725} & 0.189 & 0.948 \\
                            & AttackVLM-T   & 0.439 & 0.045 & 0.827 & 0.644 & 0.417 & 0.884 & 0.543 & 0.218 & 0.864 & 0.498 & 0.175 & 0.844 & 0.544 & 0.140 & 0.822     \\\cdashline{3-17}
                             & AttackVLM-Q   & 0.409 & 0.124  & 0.684 & 0.668  & 0.488 & 0.791 & 0.538 & 0.316 & 0.746 & 0.519 & 0.301 & 0.721 & 0.530 & 0.249 & 0.734          \\\midrule 
\multirow{3}{*}{Img2Prompt}  & Clean & \textbf{0.652} & 0.212 & 0.912 & \textbf{0.775}  & 0.454  & 0.946 & \textbf{0.699}  & 0.297 & 0.949 &  \textbf{0.684}  & 0.236 & 0.93 & \textbf{0.667} & 0.151 & 0.939 \\\cdashline{3-17}
                                & AttackVLM-T  & 0.389 & 0.097 & 0.798 & 0.626 & 0.426 & 0.866 & 0.517 & 0.214 & 0.822 & 0.481 & 0.161 & 0.797 & 0.508 & 0.129 & 0.794
          \\\cdashline{3-17}
                             & AttackVLM-Q    & 0.448 & 0.116 & 0.698 & 0.683 & 0.501  & 0.820& 0.564 & 0.316 & 0.731 & 0.536 & 0.240 & 0.761 & 0.563 & 0.270 &  0.801         \\ \bottomrule

\end{tabular}
}
\end{center}

\end{table}

\begin{table}[t]
\caption{\small{Detection accuracies. TPR is the proportion of actual adversarial images that are correctly identified. FPR is the proportion of clean images incorrectly identified as adversarial. Accuracy is the proportion of correctly identified images (both clean and adversarial).}}
\label{tab:Detectionresults}
\begin{center}
\small
\resizebox{\linewidth}{!}{
\begin{tabular}{llcccccccccccccccccc}
\toprule
\multirow{2}{*}{Victim Model}   & \multirow{2}{*}{Setting} & \multicolumn{3}{c}{RN50}     & \multicolumn{3}{c}{RN101}    & \multicolumn{3}{c}{ViT-B/16} & \multicolumn{3}{c}{ViT-B/32} & \multicolumn{3}{c}{ViT-L/14} & \multicolumn{3}{c}{Ensemble}   \\ 
                             &                                 & TPR  & FPR  & ACC            & TPR  & FPR  & ACC            & TPR     & FPR     & ACC      & TPR     & FPR     & ACC      & TPR     & FPR     & ACC      & TPR     & FPR     & ACC      \\ \midrule 
\multirow{2}{*}{UniDiffuser} & AttackVLM-T   & 0.917 & 0.085 & \textbf{0.916} & 0.912 & 0.088 & 0.912 & 0.902 & 0.098  & 0.902 & 0.368 & 0.636 & 0.366 & 0.874 & 0.127 & 0.874 & 0.87 & 0.13 & 0.87           \\\cdashline{3-20}
                             & AttackVLM-Q   & \textbf{0.925}  & 0.075 & 0.925  & 0.871 & 0.129   & 0.871 & 0.874  & 0.125  & 0.875 & 0.889  & 0.113 & 0.888 & 0.825 & 0.174  & 0.826 & 0.895 & 0.105  & 0.895             \\ \midrule 
\multirow{2}{*}{BLIP}        & AttackVLM-T       & 0.905 & 0.096 & \textbf{0.905} & 0.894 & 0.108 & 0.893 & 0.876 & 0.126 & 0.875 & 0.887 & 0.114 & 0.887& 0.84&0.159&0.841 & 0.898 &  0.103 & 0.898         \\\cdashline{3-20}
                             & AttackVLM-Q       & 0.896& 0.104 & \textbf{0.896} & 0.855 & 0.144  & 0.856 & 0.838 & 0.162 & 0.838 & 0.854 & 0.144 & 0.855 & 0.792 & 0.213 & 0.790 & 0.865 & 0.136 & 0.865            \\ \midrule 
\multirow{2}{*}{BLIP-2}      & AttackVLM-T    & 0.882 & 0.119 & 0.882 & 0.883 & 0.117 & 0.883 & 0.873& 0.128 & 0.873 & 0.898 & 0.102 & \textbf{0.898} & 0.835 & 0.166 & 0.835 & 0.891 & 0.111 &  0.890    \\\cdashline{3-20}
                             & AttackVLM-Q      & 0.921&0.082 & \textbf{0.920} & 0.885 & 0.117 & 0.884 & 0.886 & 0.114 & 0.886 & 0.896 & 0.104& 0.896 & 0.856 &  0.144 & 0.856  & 0.912  & 0.090  & 0.911  \\ \midrule 
\multirow{2}{*}{Img2Prompt}  & AttackVLM-T    & 0.841 & 0.160 & \textbf{0.841} & 0.833 & 0.170 & 0.832 & 0.815 & 0.185 & 0.815 & 0.838 & 0.164 & 0.837 & 0.783 & 0.216 & 0.784 & 0.834 & 0.167 &   0.8335     \\\cdashline{3-20}
                             & AttackVLM-Q    & 0.809 & 0.195 & \textbf{0.807} & 0.759 & 0.242 & 0.7585 & 0.767 & 0.235 & 0.766 & 0.789 & 0.213 & 0.788 & 0.708 & 0.295 & 0.707 & 0.782 & 0.220& 0.781        \\ \bottomrule

\end{tabular}
}
\end{center}

\end{table}

\begin{table}[t]
  \caption{\small{Similarity Scores for \ours{} using UniDiffuser as our T2I model and CLIP as the encoder.}}
  \label{unidiff-sim}
\begin{center}
  \setlength\tabcolsep{18pt}
\resizebox{\linewidth}{!}{\small
\begin{tabular}{llcccccc} 
\toprule
\multirow{2}{*}{Victim Model}                                     & \multirow{2}{*}{Setting} & \multicolumn{6}{c}{CLIP Image Encoder}                                         \\ 
% \cdashline{3-8}
&                                   & RN50            & RN101          & ViT-B/16       & ViT-B/32       & ViT-L/14       & Ensemble        \\ 
\midrule
\multirow{3}{*}{UniDiffuser \cite{bao2023one}}   & Clean                      & \textbf{0.737}  & \textbf{0.826} & \textbf{0.769} & 0.764          & \textbf{0.721} & \textbf{0.763}  \\ 
\cdashline{2-8}
& AttackVLM-T                      & 0.408           & 0.617          & 0.501          & \textbf{0.765} & 0.486          & 0.555           \\ 
\cdashline{2-8}
& AttackVLM-Q                         & 0.396           & 0.659          & 0.526          & 0.508          & 0.520          & 0.522           \\ 
\midrule
\multirow{3}{*}{BLIP \cite{li2022blip}}          & Clean                      & \textbf{0.713 } & \textbf{0.806} & \textbf{0.742} & \textbf{0.730} & \textbf{0.685} & \textbf{0.735}  \\ 
\cdashline{2-8}
& AttackVLM-T                      & 0.375           & 0.609          & 0.500          & 0.466          & 0.480          & 0.486           \\ 
\cdashline{2-8}
& AttackVLM-Q                         & 0.417           & 0.656          & 0.529          & 0.503          & 0.526          & 0.526           \\
% \cdashline{2-8}
% & ADV-BSA                        & 0.           & 0.          & 0.          & 0.          & 0.          & 0.           \\ 
\midrule
\multirow{3}{*}{BLIP-2 \cite{li2023blip}}        & Clean                      & \textbf{0.732}  & \textbf{0.823} & \textbf{0.764} & \textbf{0.759} & \textbf{0.720} & \textbf{0.760}  \\ 
\cdashline{2-8}
& AttackVLM-T                      & 0.425           & 0.627          & 0.533          & 0.491          & 0.517          & 0.519           \\ 
\cdashline{2-8}
& AttackVLM-Q                         & 0.390           & 0.652          & 0.511          & 0.506          & 0.510          & 0.514           \\ 
\midrule
\multirow{3}{*}{Img2Prompt \cite{10204235}}      & Clean                       & \textbf{0.663}  & \textbf{0.780} & \textbf{0.703} & \textbf{0.689} & \textbf{0.660} & \textbf{0.699}  \\ 
\cdashline{2-8}
& AttackVLM-T                      & 0.369           & 0.607          & 0.494          & 0.457          & 0.474          & 0.480           \\ 
\cdashline{2-8}
& AttackVLM-Q                         & 0.417           & 0.656          & 0.522          & 0.502          & 0.525          & 0.525           \\ 
\midrule

\multirow{2}{*}{MiniGPT-4 \cite{zhu2023minigpt}} & Clean                      & \textbf{0.599}  & \textbf{0.737} & \textbf{0.646} & \textbf{0.641} & \textbf{0.610} & \textbf{0.646}  \\ 
\cdashline{2-8}
& AttackVLM-T                      & 0.507           & 0.678          & 0.570          & 0.540          & 0.524          & 0.564           \\
\bottomrule
\end{tabular}}
\end{center}
\end{table}

%######################################################

\begin{table}[t]
\caption{\small{Detection performance for \ours{} using UniDiffuser as our T2I model and CLIP as the encoder.}}
\label{tab:DetectionresultsUD}
\begin{center}
     \setlength\tabcolsep{16pt}
\scriptsize
\resizebox{\linewidth}{!}{
\begin{tabular}{llcccccc}
\toprule
\multirow{2}{*}{Victim Model}& \multirow{2}{*}{Setting} & \multicolumn{6}{c}{CLIP Image Encoders}\\ 
% \cdashline{3-8}
 &  & RN50 & RN101 & ViT-B/16 & ViT-B/32 & ViT-L/14 & Ensemble \\ \midrule
\multirow{2}{*}{UniDiffuser \cite{bao2023one}} 
& AttackVLM-T  & \textbf{0.935}  &  0.910   & 0.910       & 0.470    & 0.910    & 0.827      \\\cdashline{2-8} 
& AttackVLM-Q   & \textbf{0.960} & 0.905    & 0.900       & 0.920    & 0.865    & 0.909       \\ \midrule
\multirow{2}{*}{BLIP \cite{li2022blip}}        
& AttackVLM-T  & 0.915  & 0.910    & 0.915       & \textbf{0.920}    & 0.845    & 0.901  \\ \cdashline{2-8} 
& AttackVLM-Q    & \textbf{0.920}  & 0.880    & 0.900       & 0.915    & 0.820    & 0.887       \\ \midrule
\multirow{2}{*}{BLIP-2 \cite{li2023blip}}      
& AttackVLM-T  & 0.915  & 0.930    & 0.885      & \textbf{0.935}    & 0.860    & 0.905      \\ \cdashline{2-8} 
& AttackVLM-Q   & \textbf{0.950}  & 0.910    & 0.920       & 0.930    & 0.860    & 0.914      \\ \midrule
\multirow{2}{*}{Img2Prompt \cite{10204235}}  
& AttackVLM-T   & \textbf{0.885}  & 0.870    & 0.830       & \textbf{0.885}    & 0.810    & 0.856    \\ \cdashline{2-8} 
& AttackVLM-Q     & \textbf{0.845}  & 0.810    & 0.805       & 0.830    & 0.775    & 0.813     \\ 

% \midrule
% \multirow{1}{*}{MiniGPT-4}      
% & AttackVLM-T   & \textbf{0.680}         &   0.620  & 0.640        & 0.635     & 0.650      & 0.645       \\ 
\bottomrule
\end{tabular}
}
\end{center}

\end{table} 

%######################################################

\begin{table}[t]
\label{tbl:CN}
  \caption{\small{Similarity Scores for \ours{} using ControlNet as our T2I model and CLIP as the encoder.}}
  \label{control-sim}
  \setlength\tabcolsep{16pt}
\scriptsize
\resizebox{\linewidth}{!}{
\begin{tabular}{llcccccc} 
\toprule
\multirow{2}{*}{Victim Model}                                   & \multirow{2}{*}{Setting} & \multicolumn{6}{c}{CLIP Image Encoder}                                         \\ 
% \cdashline{3-8}
&                                   & RN50           & RN101          & ViT-B/16        & ViT-B/32       & ViT-L/14       & Ensemble        \\ 
\midrule
\multirow{3}{*}{UniDiffuser \cite{bao2023one}} & Clean Image                       & \textbf{0.747} & \textbf{0.839} & \textbf{ 0.768} & \textbf{0.758} & \textbf{0.731} & \textbf{0.769}  \\ 
\cdashline{2-8}
& AttackVLM-T                      & 0.410          & 0.621          & 0.514           & 0.554          & 0.514          & 0.523           \\ 
\cdashline{2-8}
& AttackVLM-Q                         & 0.440          & 0.663          & 0.555           & 0.522          & 0.519          & 0.540           \\ 
\midrule
\multirow{3}{*}{BLIP \cite{li2022blip}}        & Clean Image                       & \textbf{0.747} & \textbf{0.840} & \textbf{0.770}  & \textbf{0.769} & \textbf{0.728} & \textbf{0.770}  \\ 
\cdashline{2-8}
 & AttackVLM-T                      & 0.398          & 0.625          & 0.526           & 0.494          & 0.511          & 0.511           \\ 
\cdashline{2-8}
 & AttackVLM-Q                         & 0.466          & 0.689          & 0.575           & 0.527          & 0.565          & 0.564           \\ 
\midrule
\multirow{3}{*}{BLIP-2 \cite{li2023blip}}      & Clean Image                       & \textbf{0.751} & \textbf{0.844} & \textbf{0.774}  & \textbf{0.766} & \textbf{0.735} & \textbf{0.774}  \\ 
\cdashline{2-8}
& AttackVLM-T                      & 0.388          & 0.623          & 0.526           & 0.493          & 0.512          & 0.508           \\ 
\cdashline{2-8}
& AttackVLM-Q                         & 0.463          & 0.684          & 0.571           & 0.522          & 0.565          & 0.561           \\ \midrule
\multirow{3}{*}{Img2Prompt \cite{10204235}}    & Clean Image                       & \textbf{0.661} & \textbf{0.780} & \textbf{0.712}  & \textbf{0.695} & \textbf{0.670} & \textbf{0.703}  \\ 
\cdashline{2-8}& AttackVLM-T                      & 0.400          & 0.626          & 0.532           & 0.497          & 0.514          & 0.514           \\
 & AttackVLM-Q                         & 0.463          & 0.685          & 0.569           & 0.534          & 0.569          & 0.564           \\

\bottomrule
\end{tabular}}
\end{table}

%######################################################
\begin{table}[t]
\label{tbl:DetCN}
\caption{\small{Detection Performance for \ours{} using ControlNet as our T2I model and CLIP as the encoder.}}
\label{tab:DetectionresultsCN}
\begin{center}
    \setlength\tabcolsep{16pt}
\scriptsize
\resizebox{\linewidth}{!}{
\begin{tabular}{llcccccc}
\toprule
\multirow{2}{*}{Victim Model}& \multirow{2}{*}{Setting} & \multicolumn{6}{c}{CLIP Image Encoder}\\

% \cdashline{3-8}
 &  & RN50 & RN101 & ViT-B/16 & ViT-B/32 & ViT-L/14 & Ensemble \\ \midrule
\multirow{2}{*}{UniDiffuser \cite{bao2023one}} 
& AttackVLM-T  & 0.935  &   \textbf{0.980}   & 0.925       & 0.895    & 0.920    & 0.931      \\\cdashline{2-8} 
& AttackVLM-Q   & 0.945  &   \textbf{0.965}   & 0.880       & 0.920     & 0.880   & 0.918       \\ \midrule
\multirow{2}{*}{BLIP \cite{li2022blip}}        
& AttackVLM-T  & 0.955 & \textbf{0.965} & 0.880 & 0.945 & 0.880 & 0.925 \\ \cdashline{2-8} 
& AttackVLM-Q    & \textbf{0.940} & 0.905 & 0.870 &  0.925 & 0.830 &  0.894       \\ \midrule
\multirow{2}{*}{BLIP-2 \cite{li2023blip}}      
& AttackVLM-T   & 0.935        &   \textbf{0.950}  & 0.905        & 0.930     & 0.900     & 0.924       \\ \cdashline{2-8} 
& AttackVLM-Q    & \textbf{0.915} & 0.910    & 0.880     & 0.890   &   0.850  & 0.889      \\ 
\midrule
\multirow{2}{*}{Img2Prompt \cite{10204235}}  
& AttackVLM-T   & \textbf{0.965}  &   0.940  & 0.900       & 0.950    & 0.890    & 0.929      \\ \cdashline{2-8} 
& AttackVLM-Q      & \textbf{0.950}  &   0.895   & 0.860       &  0.915    & 0.800    & 0.884      \\ 

% \multirow{1}{*}{MiniGPT-4}      
% & AttackVLM-T   & \textbf{0.680}         &   0.620  & 0.640        & 0.635     & 0.650      & 0.645       \\ 
\bottomrule
\end{tabular}
}
\end{center}

\end{table} 

\begin{algorithm}[t]
\small{
\caption{Adaptive Attack using Learnable Adapters}
\label{adaptiveAlgo}
\begin{algorithmic}[1]
    \State \textbf{Input:} Original image $x_{in}$, target caption $t$
    \State \textbf{Output:} Adversarial image $x_{\text{adv}}$
    \State \textbf{Initialize:} $\delta \leftarrow 0$
    \State \textbf{VLM Model:} $\mathcal{F}_{\theta}(x_{\text{in}}; p) \rightarrow t$ \Comment{Victim model generates caption}
    \State \textbf{VLM Text Encoder:} $\hat{\mathcal{F}}_{\theta}(x_{\text{in}}) \rightarrow z$
    \State \textbf{T2I Image Generator:} $\hat{G}_{\psi}(z) \rightarrow x_{gen}$
    \State \textbf{Adapter Network Training:} Train adapter $\mathcal{A}$ 
    \Repeat
        \State $x_{adv} \leftarrow x_{in} + \delta$
        \State $\; z \leftarrow \hat{\mathcal{F}}_{\theta}(x_{adv})$
        \State $\; z' \leftarrow \mathcal{A}(z)$
        \State $\; x_{gen} \leftarrow \hat{G}_{\psi}(z')$
        \For{$j = 1$ to $N$}
            \State $v^{(j)}_{\text{adv}} \leftarrow \mathcal{I}_{\phi j, \xi}(x_{adv})$
            \State $v^{(j)}_{\text{gen}} \leftarrow \mathcal{I}_{\phi j, \xi}(x_{gen})$
        \EndFor
        \State \textbf{Compute broken-down loss terms:}
        \State \quad $L_{\text{img-target}} \leftarrow d\!\big(\tilde{\mathcal{I}}_\phi(x_{adv}),\, \tilde{\mathcal{I}}_\phi(G_\psi(t^*; \eta))\big)$
        \State \quad $L_{\text{adv-gen}} \leftarrow \tfrac{1}{N}\sum_{j=1}^N d\!\big(v^{(j)}_{\text{adv}},\, v^{(j)}_{\text{gen}}\big)$
        \State \quad $L_{\text{total}} \leftarrow L_{\text{img-target}} + L_{\text{adv-gen}}$
        \State \textbf{Update $\delta$:}
        \State \quad $\delta \leftarrow \delta - \gamma \cdot \nabla_{\delta} L_{\text{total}}$
    \Until {Convergence}
    \State $x_{\text{adv}} \leftarrow x_{in} + \delta$
    \State \Return $x_{\text{adv}}$
\end{algorithmic}}
\end{algorithm}

\paragraph{Robustness to Adaptive Attacks.}
\label{adaptappendix}

To further evaluate the robustness of \ours{} against adaptive adversaries, we introduce an additional adaptive attack similar to the BPDA+EOT attack discussed in the main paper, but with a different optimization objective. This attack constructs a fully differentiable end-to-end pipeline by linking the victim vision-language model (VLM) and a text-to-image (T2I) generative model through a learned adapter network. The attacker’s objective is to generate an adversarial image $x_{adv}$ that simultaneously aligns with the target caption $t^*$ and its corresponding generated image $x_{gen}$, thereby attempting to bypass our detection mechanism. The optimization objective (\cref{adaptiveAlgo}) enforces similarity between these representations while employing randomized encoders and Expectation over Transformation (EOT) to approximate gradients through the defense. We assess the attacker’s success under varying levels of knowledge about the image encoders used in \ours{}. \\

\noindent As shown in \cref{adaptiveAblation}, detection performance improves as the number of encoders increases and when stochasticity (OTU approach) is applied, significantly hindering the attacker’s ability to evade detection. Even under the strongest assumption, where the attacker knows all encoders, detection accuracy remains high. Moreover, we also compared text-embedding similarities between target and generated captions for standard (ADV-Transfer) and adaptive attacks \cref{clip_text}. The adaptive attacks yield notably lower similarity scores, demonstrating reduced attack success. These results confirm that \ours{} maintains strong robustness even against highly adaptive, gradient-based attacks, benefiting from encoder diversity and stochastic defense components.

\begin{table}[t]
\caption{\small{Robustness of \ours{} on adversarial samples generated through adaptive attacks based on the attacker's knowledge of image encoders used in \ours{}. The defender employs between one and five pretrained CLIP image encoders with backbones RN50, RN101, ViT-B/16, ViT-B/32, and ViT-L/14. The attacker has knowledge of all, all but one, or all but two of these encoders, and replaces unknown ones with OpenCLIP encoders.}}
\label{adaptiveAblation}
\begin{center}
\scriptsize
\resizebox{\linewidth}{!}{
\begin{tabular}{lcccccc}
\hline
\multirow{2}{*}{Attacked Image Encoder} & \multicolumn{3}{c}{\ours{}} & \multicolumn{3}{c}{\ours{} (OTU approach)} \\ 
& ALL & ALL but ONE & ALL but TWO & ALL & ALL but ONE & ALL but TWO \\ \hline
ViT-B/32 & 0.55 & 0.90 & - & 0.50 & 0.90 & - \\\cdashline{1-7}
RN50 and ViT-B/32 & 0.60 & 0.70 & 0.90 & 0.70 & 0.80 & 0.90 \\ \cdashline{1-7}
RN50, ViT-B/32, and ViT-L/14 & 0.65 & 0.65 & 0.80 & 0.75 & 0.75 & 0.80 \\ \cdashline{1-7}
RN50, ViT-B/16, ViT-B/32, and ViT-L/14 & 0.65 & 0.65 & 0.85 & 0.75 & 0.80 & 0.85 \\ \cdashline{1-7}
RN50, RN101, ViT-B/16, ViT-B/32, and ViT-L/14 & 0.75 & 0.75 & 0.85 & 0.85 & 0.90 & 0.80 \\ 
\hline
\end{tabular}
}
\end{center}
\end{table}

\begin{table}[ht!]
\caption{\small{Text embedding similarity between the target captions and captions produced by the victim model under transfer (ADV-Transfer) and adaptive attacks. Lower similarity indicates stronger robustness.}}
\label{clip_text}
\begin{center}
\scriptsize
\resizebox{\linewidth}{!}{
\begin{tabular}{llcccccc} 
\toprule
\multirow{2}{*}{Victim Model} & \multirow{2}{*}{Setting} & \multicolumn{6}{c}{CLIP Image Encoder} \\ 
 & & RN50 & RN101 & ViT-B/16 & ViT-B/32 & ViT-L/14 & Ensemble  \\ 
\midrule
\multirow{6}{*}{UniDiffuser} & ADV-Transfer & 0.76 & 0.71 & 0.74 & 0.77 & 0.68 & 0.73 \\ 
\cdashline{2-8}
 & Adaptive (ViT-B/32) & 0.59 & 0.61 & 0.60 & 0.64 & 0.53 & 0.60 \\ 
\cdashline{2-8}
 & Adaptive (RN50 + ViT-B/32) & 0.63 & 0.60 & 0.63 & 0.66 & 0.55 & 0.61 \\ 
\cdashline{2-8}
 & Adaptive (RN50 + ViT-B/32 + ViT-L/14) & 0.70 & 0.64 & 0.68 & 0.70 & 0.60 & 0.66 \\ 
\cdashline{2-8}
 & Adaptive (RN50 + ViT-B/16 + ViT-B/32 + ViT-L/14) & 0.69 & 0.64 & 0.68 & 0.71 & 0.62 & 0.67 \\ 
\cdashline{2-8}
 & Adaptive (RN50 + RN101 + ViT-B/16 + ViT-B/32 + ViT-L/14) & 0.63 & 0.64 & 0.66 & 0.67 & 0.58 & 0.64 \\ 
\bottomrule
\end{tabular}
}
\end{center}
\end{table}

% *****************************************************************************************

\begin{figure}[ht]
    \centering
    \includegraphics[width=\linewidth]{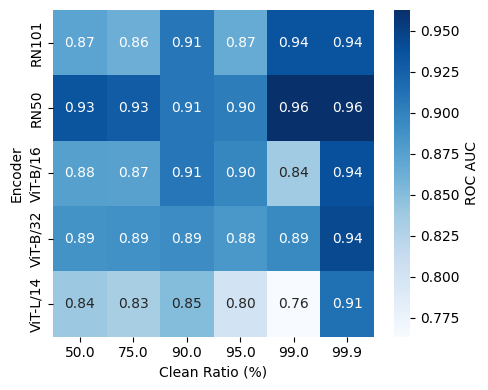}
    \vspace{-25pt}
    \caption{Effect of Clean Ratio on Detection Accuracy across Different Encoders.}
    \label{cleanratio}
    \vspace{-10pt}
\end{figure}

% \newpage
\paragraph{Impact of Clean Ratio on Detection Accuracy.} \cref{cleanratio} illustrates how varying the proportion of clean to adversarial examples affects detection accuracy. As the clean ratio increases from 50\% to 99.9\%, overall performance consistently improves. This trend is most evident for the RN50 encoder, which maintains strong ROC AUC scores even at lower clean ratios. In contrast, encoders such as ViT-L/14 are more sensitive to reduced clean ratios, exhibiting a noticeable performance drop, particularly near the 99\% level. These observations indicate that some encoders are inherently more robust to imbalanced datasets. Our ensemble approach effectively mitigates these disparities, combining the strengths of different encoders to deliver stable, well-rounded performance. Notably, detection performance stabilizes at the highest clean ratio (99.9\%), where all encoders achieve their best or near-best results. Overall, these findings demonstrate that our method remains reliable and accurate across a wide range of clean-to-adversarial distributions, even when adversarial interference is minimal.

% %######################################################

% \clearpage
\section{Qualitative Examples}
\label{vizz}
\begin{figure}[ht]
    \centering

    % \subfloat[]
    {\includegraphics[width=1\linewidth]{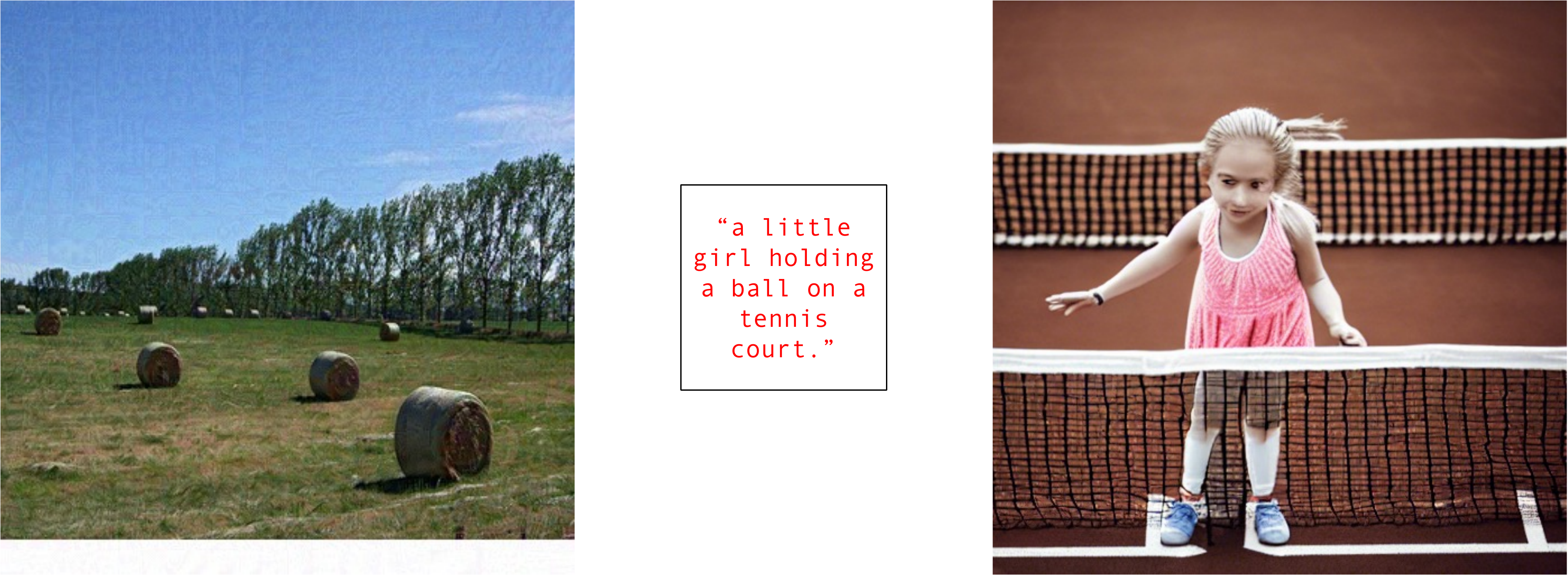}} 
    {\includegraphics[width=1\linewidth]{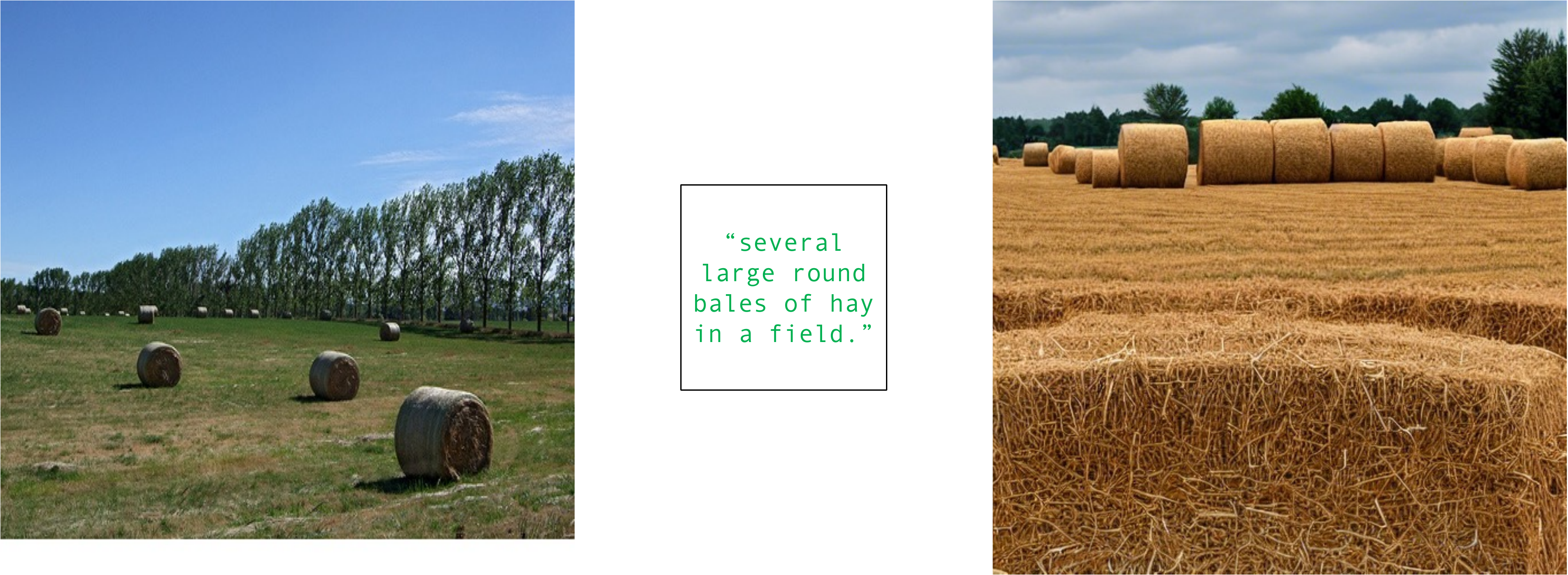}}\\
    \vspace{4em}
    
    {\includegraphics[width=1\linewidth]{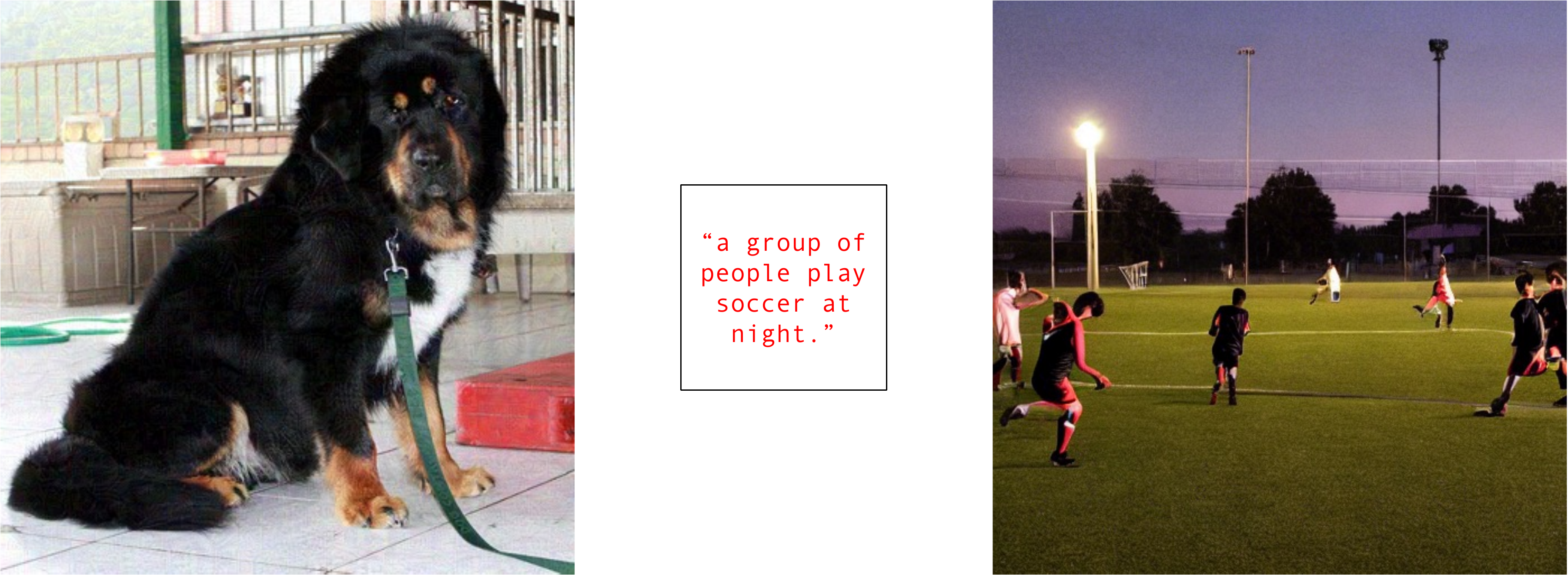}} 
    {\includegraphics[width=1\linewidth]{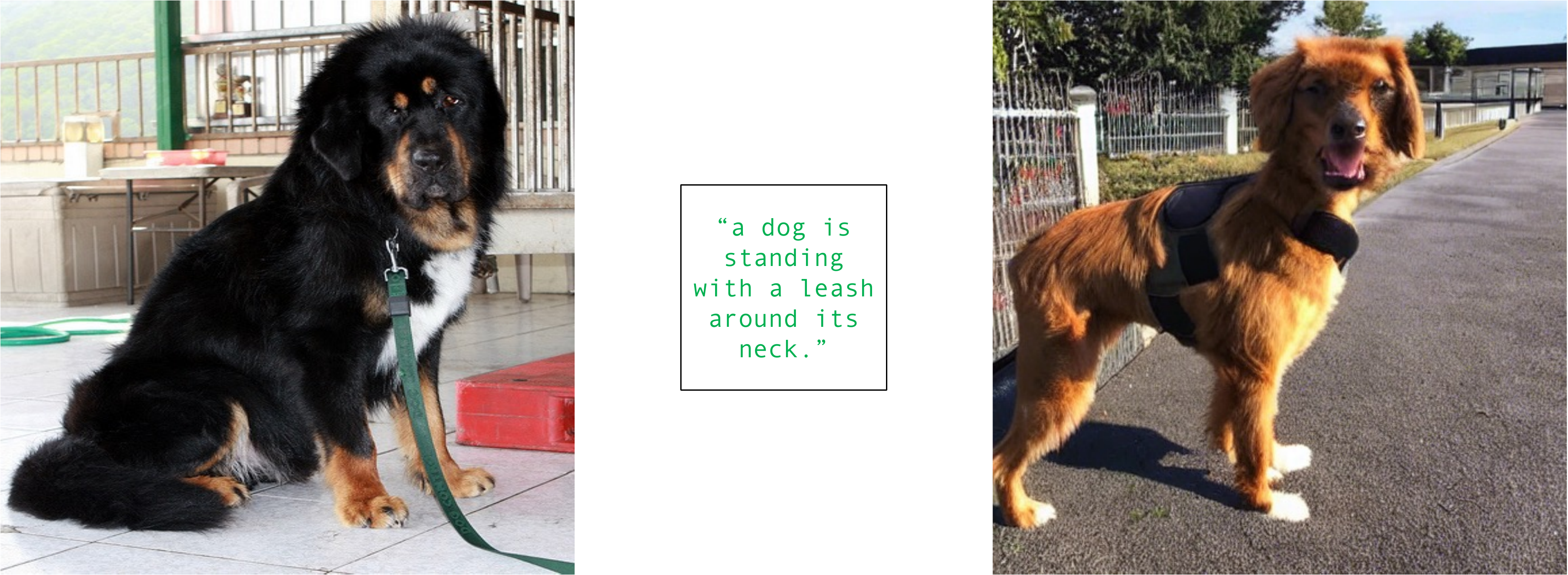}} \\
    \vspace{4em}
    {\includegraphics[width=1\linewidth]{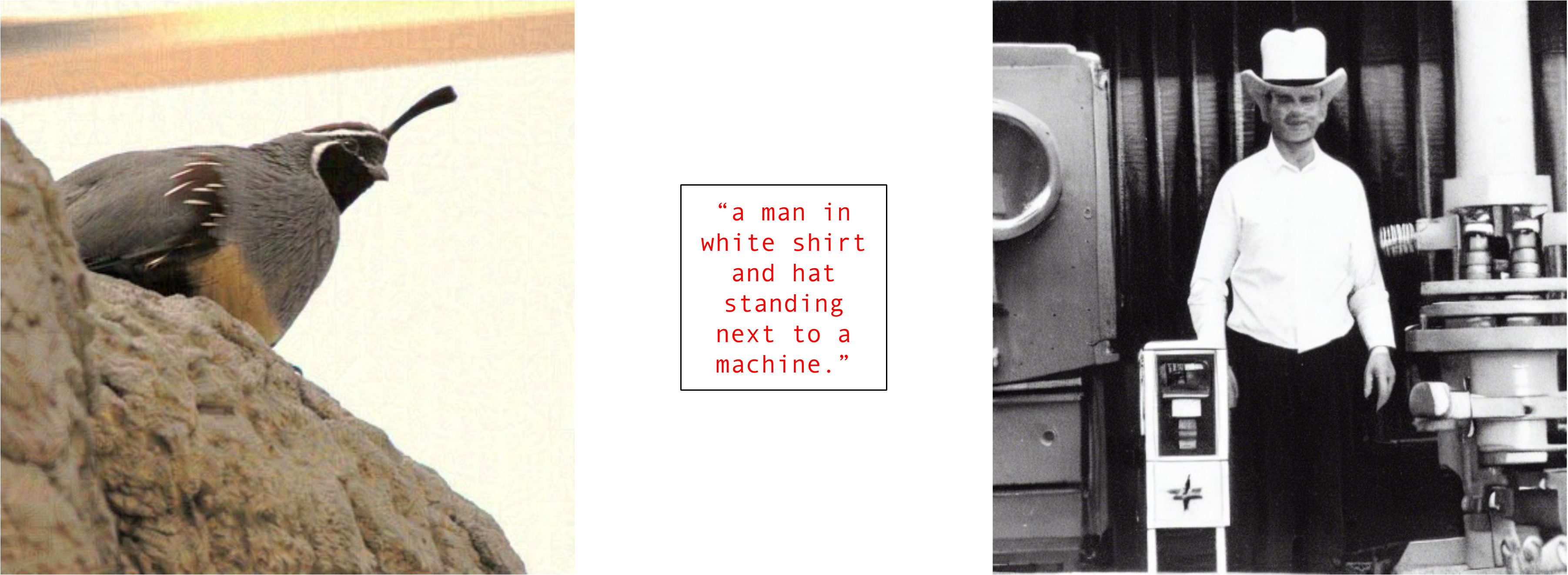}} 
    {\includegraphics[width=1\linewidth]{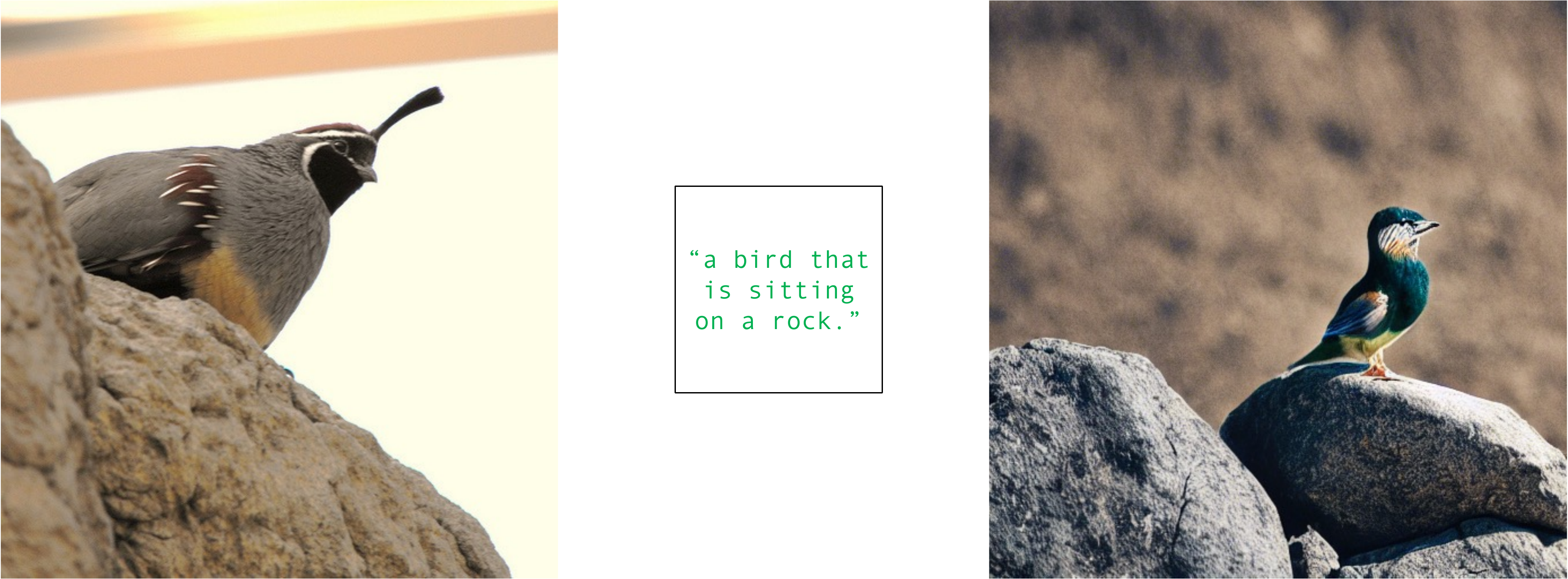}}

    \caption{\small{Visual results using BLIP (Victim Model) and Stable Diffusion (T2I Model).}}
\end{figure}

\begin{figure}[t]
    \centering

    {\includegraphics[width=1\linewidth]{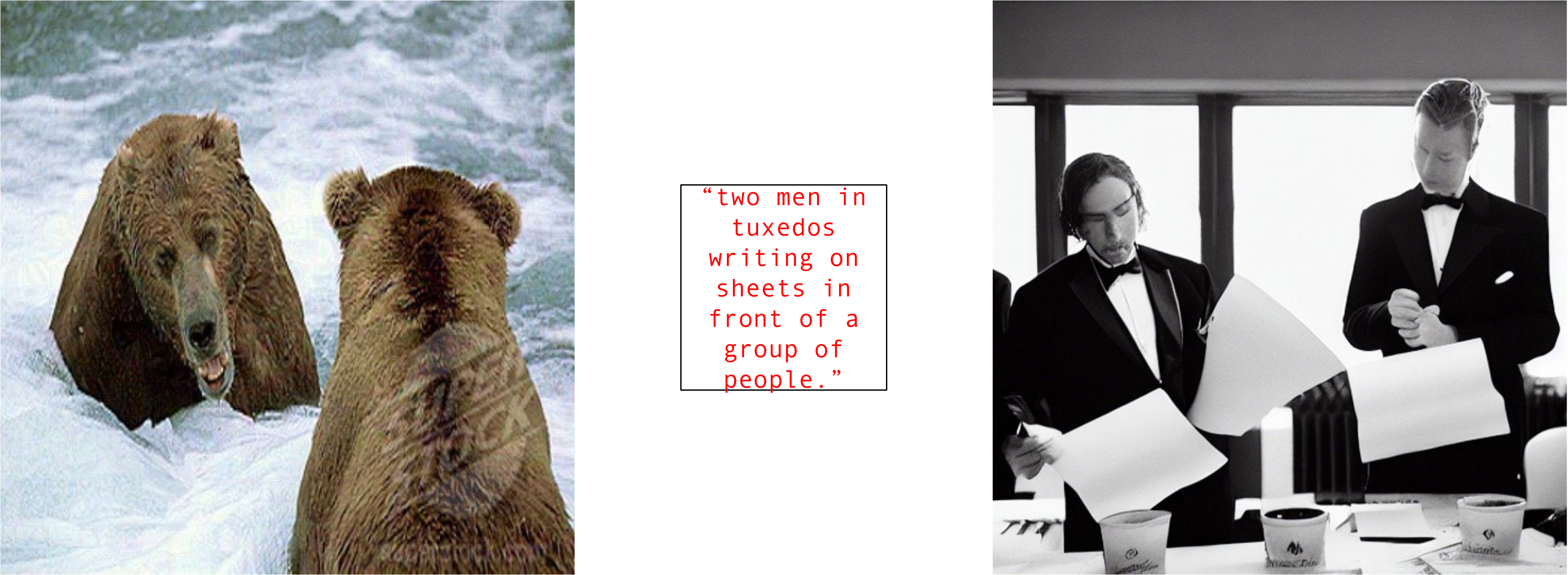}}
    {\includegraphics[width=1\linewidth]{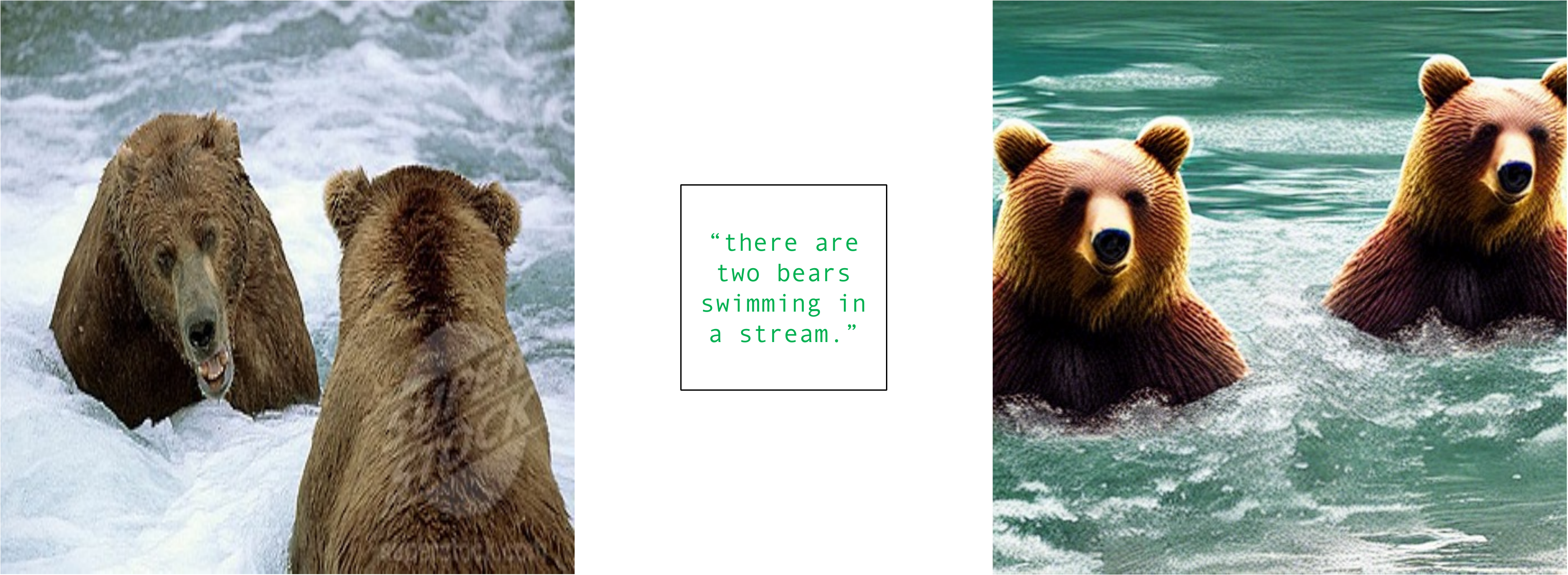}} \\
    \vspace{4em}
    {\includegraphics[width=1\linewidth]{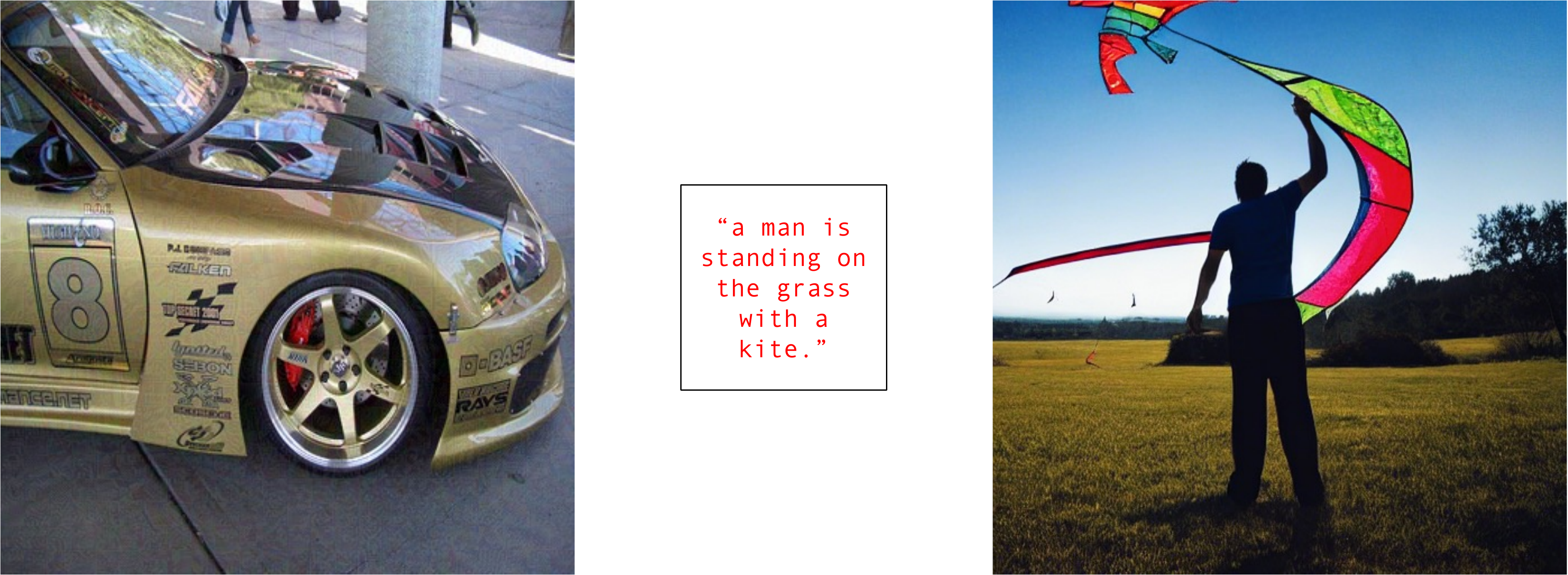}}
    {\includegraphics[width=1\linewidth]{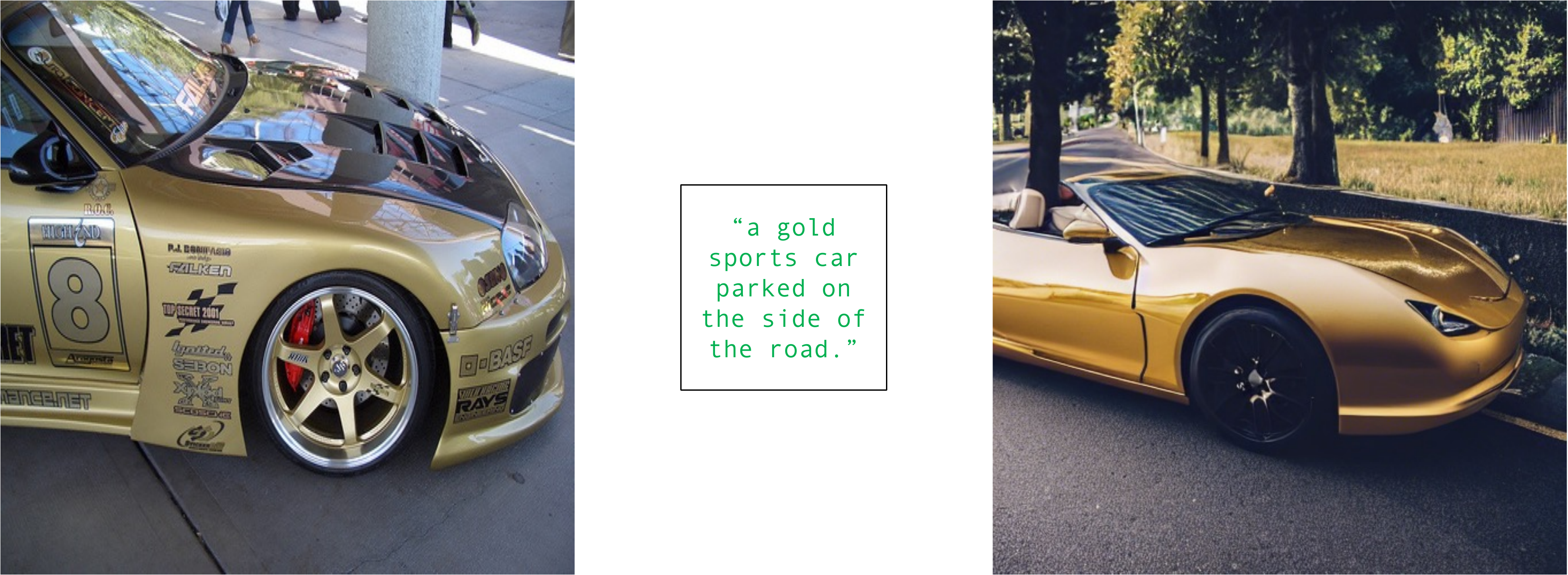}} \\
    \vspace{4em}
    {\includegraphics[width=1\linewidth]{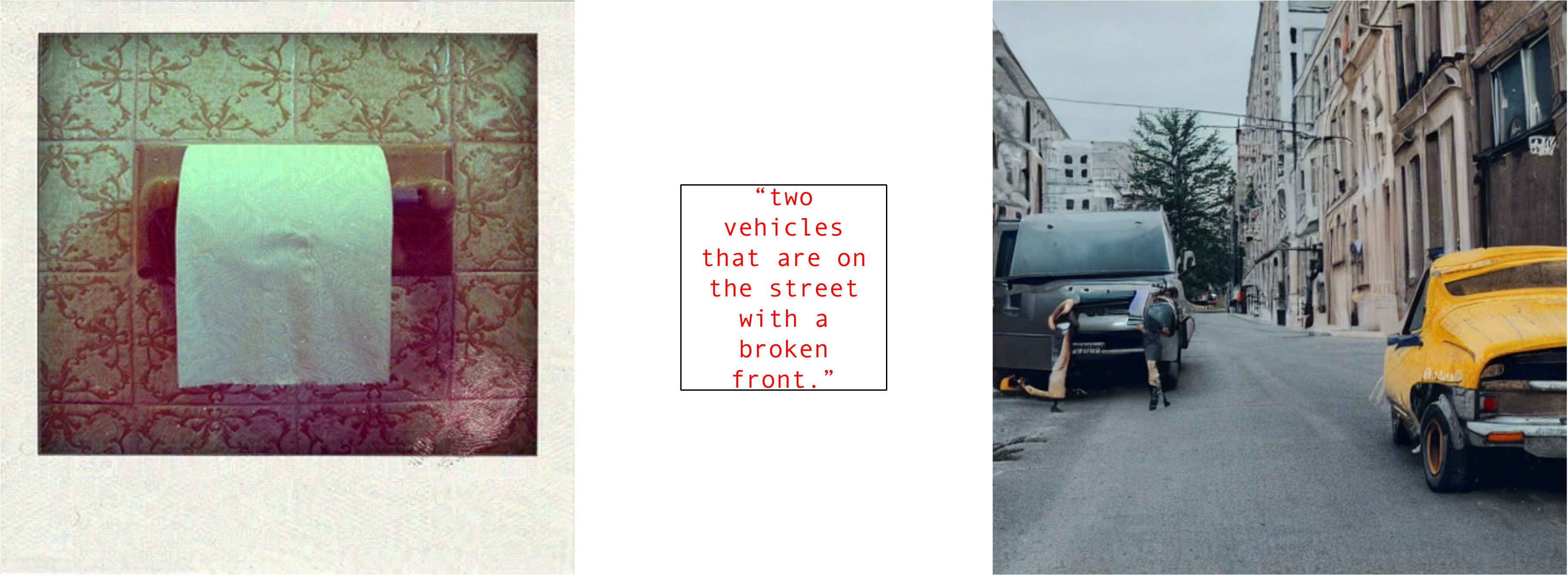}}
    {\includegraphics[width=1\linewidth]{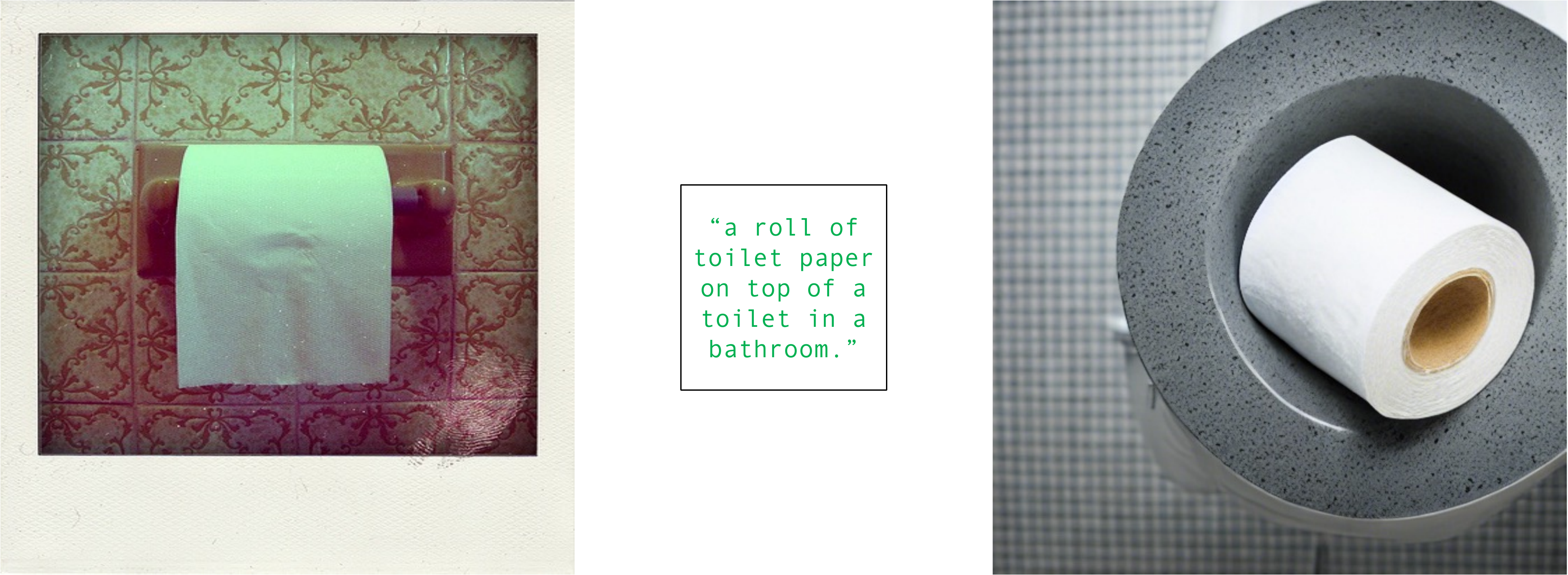}}

    \caption{\small{Visual results using BLIP (Victim Model) and Stable Diffusion (T2I Model).}}
\end{figure}

\begin{figure}[t]
    \centering

    {\includegraphics[width=1\linewidth]{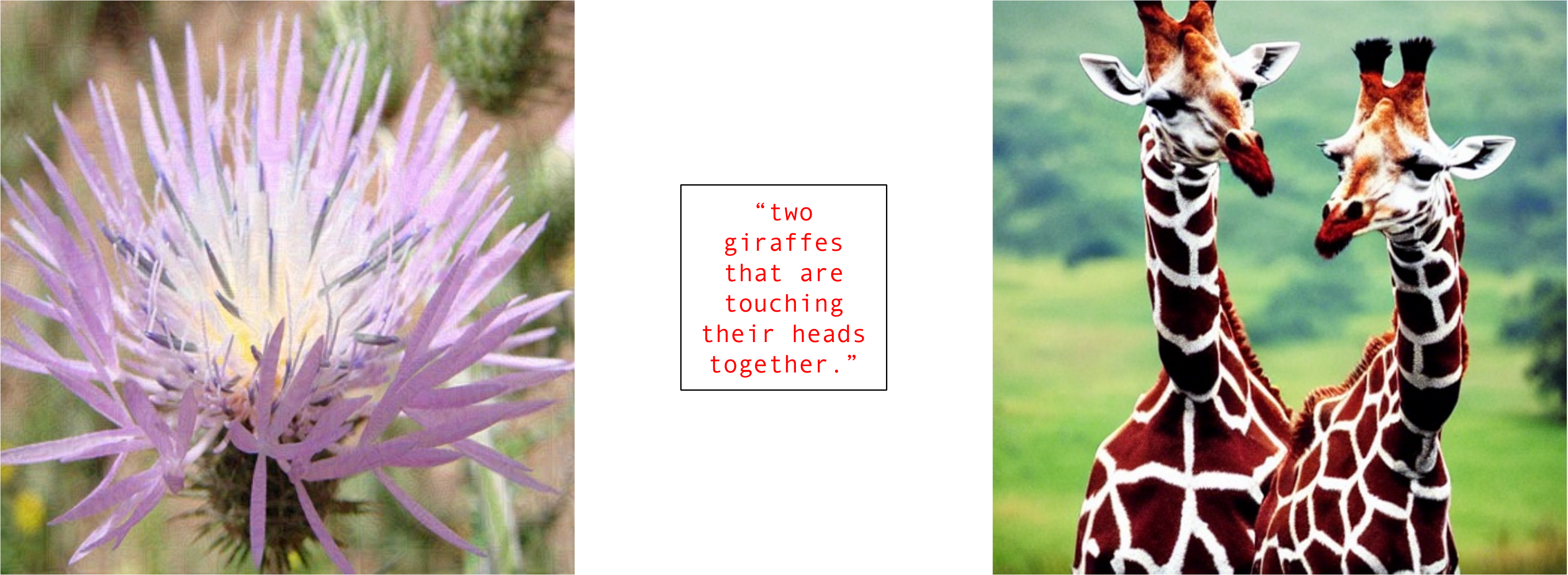}}
    {\includegraphics[width=1\linewidth]{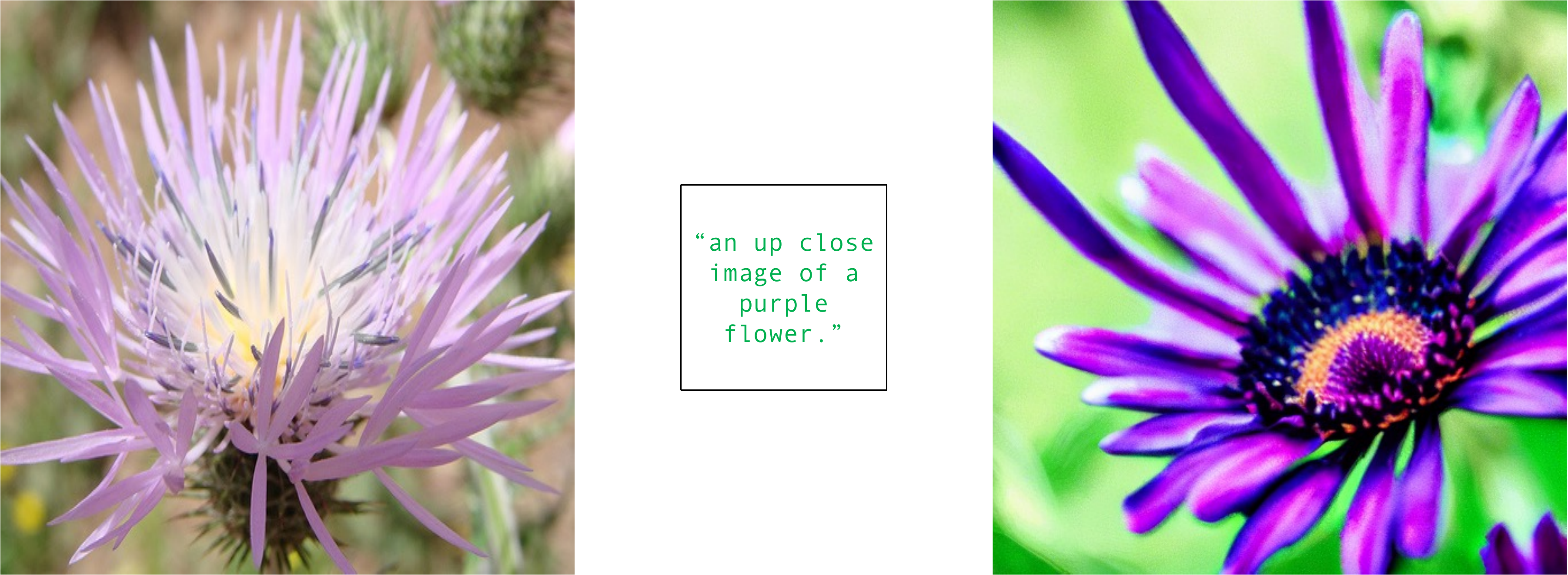}} \\
    \vspace{4em}
    {\includegraphics[width=1\linewidth]{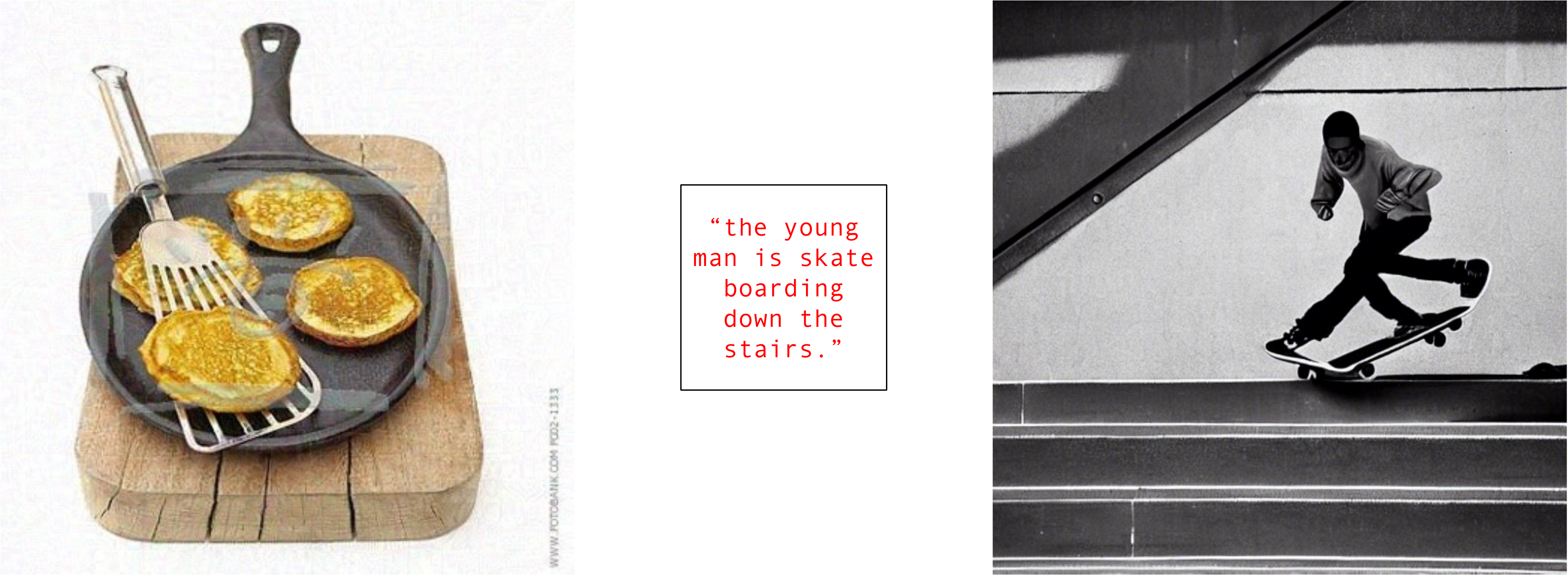}}
    {\includegraphics[width=1\linewidth]{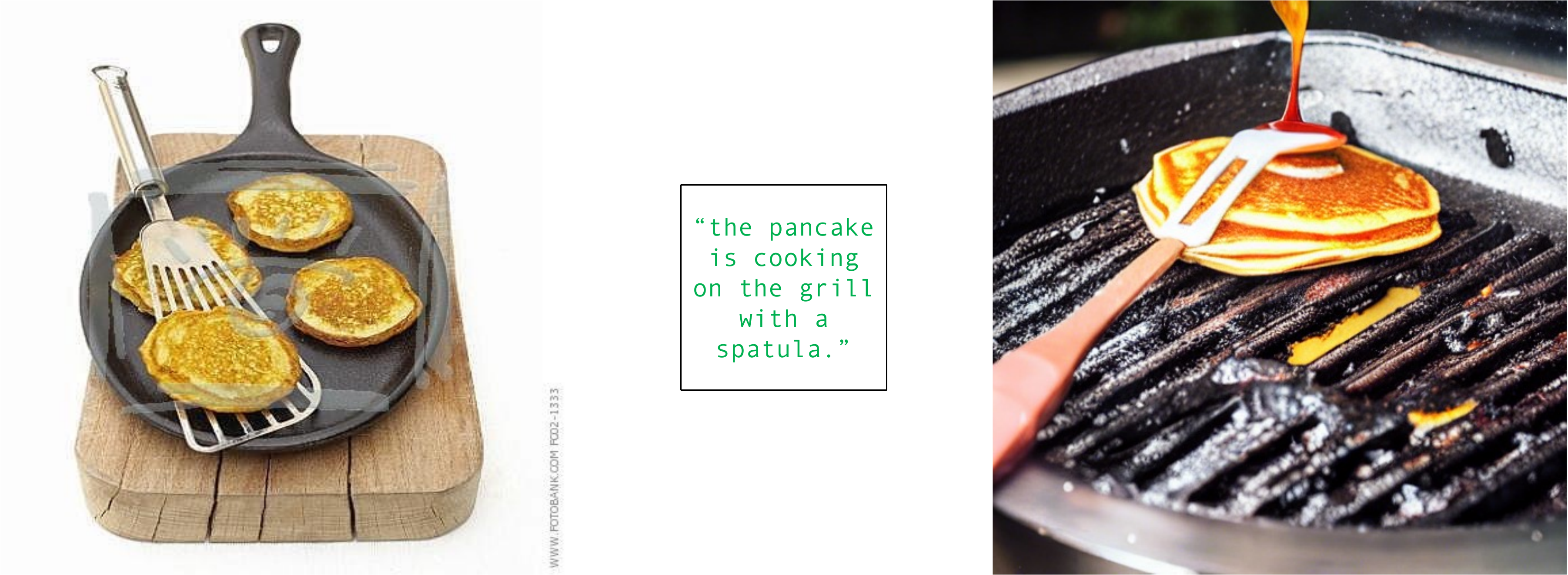}} \\
    \vspace{4em}
    {\includegraphics[width=1\linewidth]{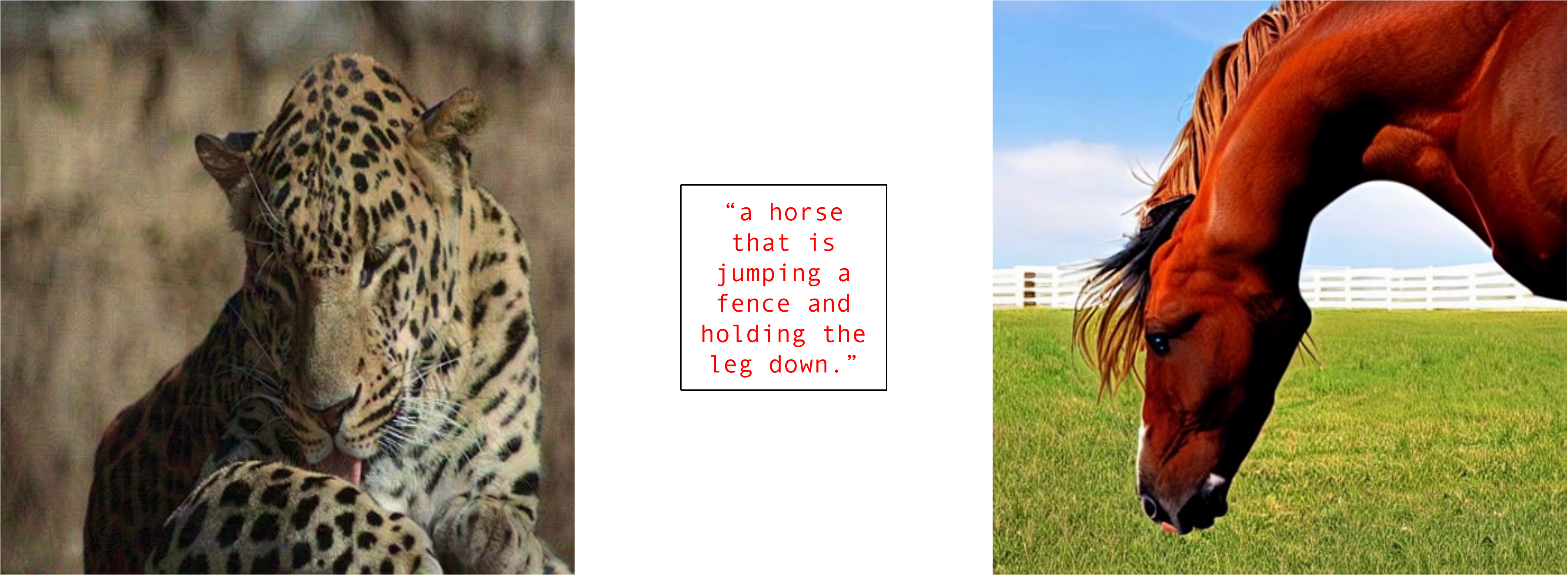}}
    {\includegraphics[width=1\linewidth]{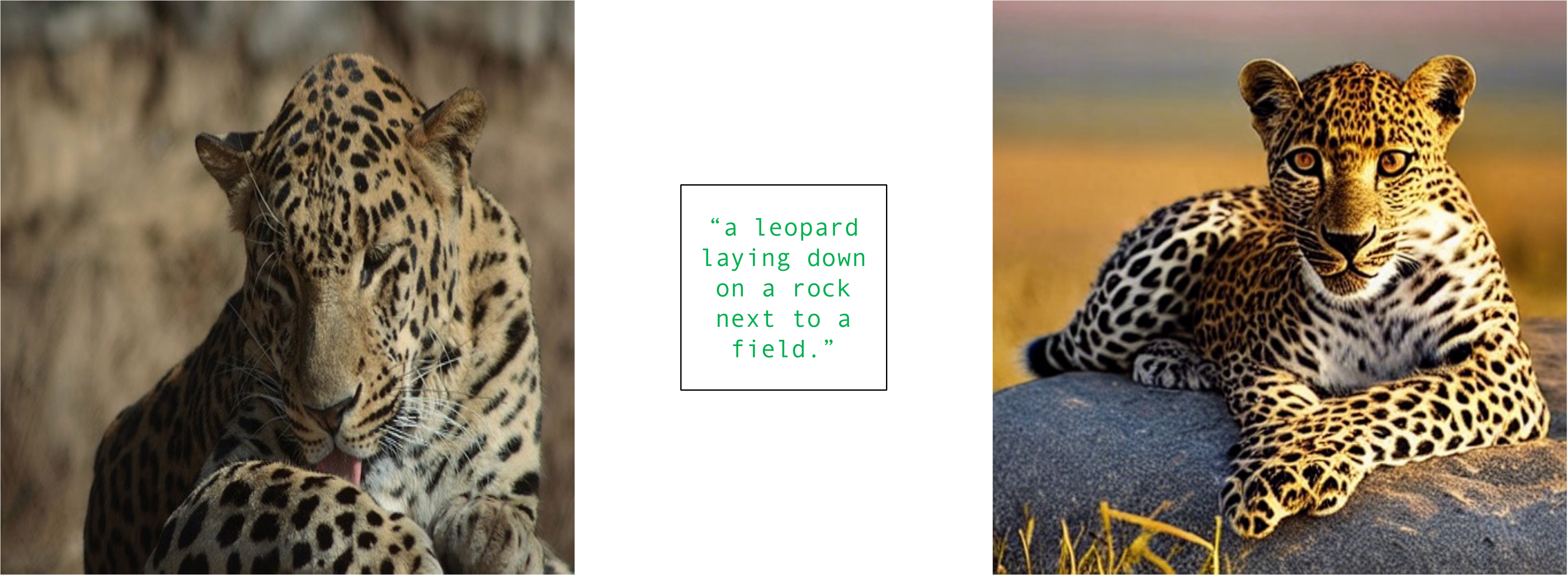}}

    \caption{\small{Visual results using BLIP (Victim Model) and Stable Diffusion (T2I Model).}}
\end{figure}

% \clearpage

\end{document}